\documentclass[runningheads]{llncs}
\usepackage{hyperref}
\usepackage{graphicx}
\usepackage{amsmath,amssymb} %

\DeclareMathOperator*{\argmin}{arg\,min}
\usepackage{color}
\usepackage[linesnumbered,ruled]{algorithm2e}
\usepackage{caption}
\usepackage{subcaption}
\usepackage{rotating}
\usepackage{multirow}
\usepackage{pgf,tikz}
\usepackage{pgfplots}
\usepackage{dsfont}
\usepackage{ multicol, latexsym}
\usepackage{float}
\usepackage[width=122mm,left=12mm,paperwidth=146mm,height=193mm,top=12mm,paperheight=217mm]{geometry}
\begin{document}
\newcommand{\bryan}[1]{\textcolor{green}{Bryan: #1}}
\newcommand{\matt}[1]{\textcolor{red}{Matt: #1}}
\definecolor{aqua}{rgb}{0.0, 0.48, 0.65}

\newcommand{\thibault}[1]{\textcolor{aqua}{Thibault: #1}}

\newcommand{\mathieu}[1]{\textcolor{blue}{Mathieu: #1}}
\newcommand{\vk}[1]{\textcolor{magenta}{Vova: #1}}
\newcommand{\mf}[1]{\textcolor{green}{[Matt: #1]}}
\newcommand{\todo}[1]{\textcolor{red}{TODO: #1}}
\newcommand{\remove}[1]{\textcolor{red}{\sout{#1}}}
\newcommand{\ournet}{AtlasNet}
\newcommand{\patch}{learnable parametrization} %
\newcommand{\patches}{learnable parametrizations}
\newcommand{\atlas}{learnable atlas}
\newcommand{\myparagraph}[1]{\noindent{\bf #1}}

\pagestyle{headings}
\mainmatter
\def\ECCV18SubNumber{1804}  %
\title{
3D-CODED : 3D Correspondences by Deep Deformation 
} %
\titlerunning{
3D-CODED : 3D Correspondences by Deep Deformation 
} %

\authorrunning{T. Groueix, M. Fisher, V. G. Kim, B. C. Russell, M. Aubry}

\author{
Thibault Groueix\inst{1}\and Matthew Fisher\inst{2}\and Vladimir G. Kim\inst{2}\and Bryan C. Russell\inst{2}\and Mathieu Aubry\inst{1}}
\institute{LIGM (UMR 8049), \'Ecole des Ponts, UPE \and Adobe Research \\ {\tt\small \url{http://imagine.enpc.fr/\~groueixt/3D-CODED/}}\\
}

\maketitle

\begin{abstract}
We present a new deep learning approach for matching deformable shapes by introducing {\it Shape Deformation Networks} which jointly encode 3D shapes and correspondences. This is achieved by factoring the surface representation into (i) a template, that parameterizes the surface, and (ii) a learnt global feature vector that parameterizes the %
transformation of the template into the input surface. %
By predicting this feature for a new shape, we implicitly predict correspondences between this shape and the template. We show that these correspondences can be improved by an additional step which improves the shape feature by minimizing the Chamfer distance between the input and transformed template. %
We demonstrate that our simple approach improves on state-of-the-art results on the difficult FAUST-inter challenge, with an average correspondence error of 2.88cm. 
We show, on the TOSCA dataset, that our method is robust to many types of perturbations, and generalizes to non-human shapes. This robustness allows it to perform well on real unclean, meshes from the the SCAPE dataset. %
\end{abstract}
\begin{figure}[h]
\centering
\begin{subfigure}[b]{\linewidth}
\includegraphics[width=0.9\linewidth]{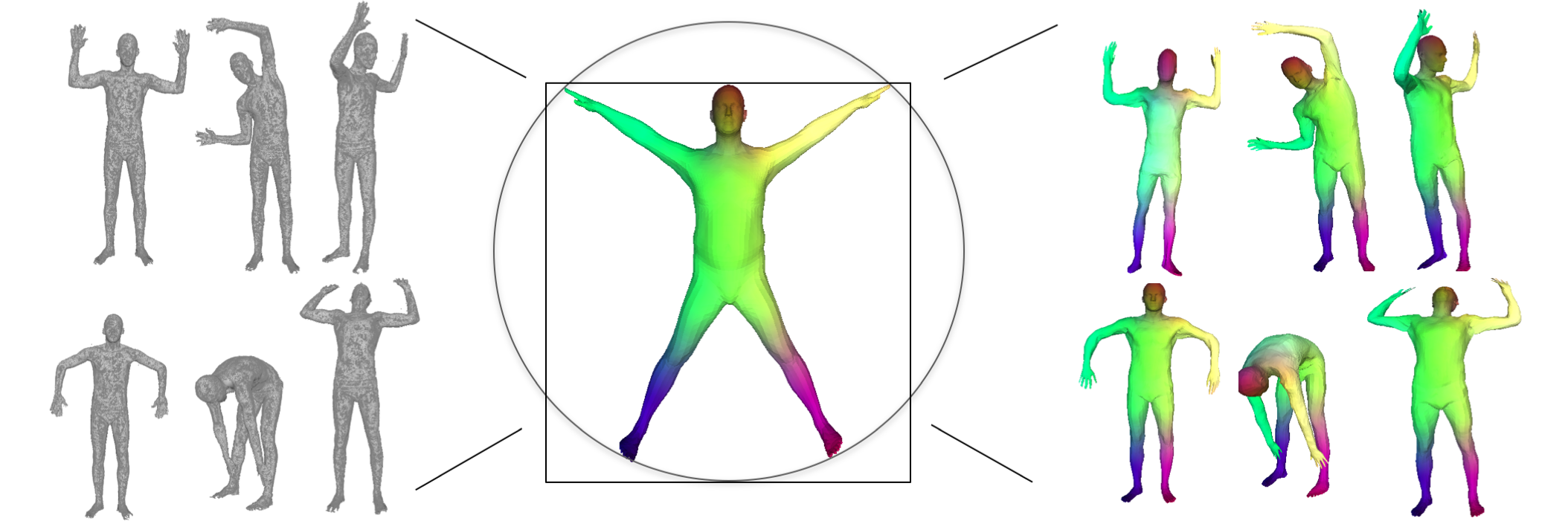}
\end{subfigure}\\
\begin{subfigure}[b]{0.3\linewidth}
 \caption{\bf Input Shape}
\end{subfigure}
\begin{subfigure}[b]{0.2\linewidth}
 \caption{\bf Template}
\end{subfigure}
\begin{subfigure}[b]{0.45\linewidth}
 \caption{\bf Deformed template}
\end{subfigure}
\caption{Our approach predicts shape correspondences by learning a consistent mesh parameterization with a shared template. Colors show correspondences.}
\end{figure}

\section{Introduction}
There is a growing demand for techniques that make use of the large amount of 3D content generated by modern sensor technology. An essential task is to establish reliable 3D shape correspondences between scans from raw sensor data or between scans and a template 3D shape. This process is challenging due to low sensor resolution and high sensor noise, especially for articulated shapes, such as humans or animals, that exhibit significant non-rigid deformations and shape variations. %

Traditional approaches to estimating shape correspondences for articulated objects typically rely on intrinsic surface analysis either optimizing for an isometric map or leveraging intrinsic point descriptors~\cite{Sun10}. To improve correspondence quality, these methods have been extended to take advantage of category-specific data priors~\cite{add16}. Effective human-specific templates and registration techniques have been developed over the last decade~\cite{Zuffi15}, but these methods require significant effort and domain-specific knowledge to design the parametric deformable template, create an objective function that
ensures alignment of salient regions and is not prone to being stuck in local minima, and develop an optimization strategy that effectively combines a global search for a good heuristic initialization and a local refinement procedure.

In this work, we propose {\em Shape Deformation Networks}, a comprehensive, all-in-one solution to template-driven shape matching. 
A Shape Deformation Network learns to deform a template shape to align with an input observed shape. %
Given two input shapes, we align the template to both inputs and obtain the final map between the inputs by reading off the correspondences from the template. 

We train our Shape Deformation Network as part of an encoder-decoder architecture, which jointly learns an encoder network that takes a target shape as input and generates a global feature representation, and a decoder Shape Deformation Network that takes as input the global feature and deform the template into the target shape. At test time, we improve our template-input shape alignment by optimizing locally the Chamfer distance between target and generated shape over the global feature representation which is passed in as input to the Shape Deformation Network. %
Critical to the success of our Shape Deformation Network is the ability to learn to deform a template shape to targets with varied appearances and articulation. 
We achieve this ability by training our network on a very large corpus of shapes. %

In contrast to previous work~\cite{Zuffi15}, our method does not require a manually designed deformable template; the deformation parameters and degrees of freedom are implicitly learned by the encoder. 
Furthermore, while our network can take advantage of known correspondences between the template and the example shapes, which are typically available when they have been generated using some parametric model ~\cite{Bogo:CVPR:2014,varol17a}, we show it can also be trained without correspondence supervision. 
This ability allows the network to learn from a large collection of shapes lacking explicit correspondences.

We demonstrate that with sufficient training data this simple approach achieves state-of-the-art results and outperforms techniques that require complex multi-term objective functions instead of the simple reconstruction loss used by our method. %

\section{Related work}

Registration of non-rigid geometries with pose and shape variations is a 
long standing problem with extensive prior work. We first provide a brief overview of 
generic correspondence techniques. %
We then focus on category specific and template matching methods developed
for human bodies, which are more closely related to our approach. Finally, we present an overview of deep learning approaches that have been developed for shape matching and more generally for working with 3D data.

\subsubsection{Generic shape matching.} To estimate correspondence between articulated objects, it is common to assume that 
their intrinsic structure (e.g., geodesic distances) remains relatively consistent across
all poses~\cite{Memoli05}. 
Finding point-to-point correspondences that minimize metric distortion is a non-convex 
optimization problem, referred to as generalized multi-dimensional scaling~\cite{Bronstein06gmds}. 
This optimization is typically sensitive to an initial guess~\cite{Bronstein06}, and thus existing
techniques rely on local feature point descriptors such as HKS~\cite{Sun10} and WKS~\cite{Aubry11}, 
and use hierarchical optimization strategies~\cite{Sahillioglu11,Raviv13}.
Some relaxations of this problem have been proposed such as: 
formulating it as Markov random field and using linear programming relaxation~\cite{Chen15}, 
optimizing for soft correspondence~\cite{Solomon12,Kim12,Solomon16}, restricting  
correspondence space to conformal maps~\cite{Lipman09,Kim11}, heat kernel maps~\cite{Ovsjanikov10}, 
and aligning functional bases~\cite{Ovsjanikov12}. 

While these techniques are powerful generic tools, some common categories, such as humans, 
can benefit from a plethora of existing data~\cite{Bogo:CVPR:2014} to leverage stronger 
class-specific priors.

\subsubsection{Template-based shape matching.}  A natural way to leverage class-specific knowledge is through the explicit use of a shape model. While such template-based techniques provide the best correspondence results they require a careful 
parameterization of the template, which took more than a decade of research to reach the current 
level of maturity~\cite{Allen02,Allen03,Allen06,Loper15,Zuffi15}. For all of these techniques, 
fitting this representation to an input 3D shape requires also designing an objective function
that is typically non-convex and involves multiple terms to guide the optimization to the 
right global minima. 
In contrast, our method only relies on a single template 3D mesh and surface reconstruction
loss. It leverages a neural network to learn how to parameterize the human body while optimizing
for the best reconstruction. %

\subsubsection{Deep learning for shape matching.} 
Another way to leverage priors and training data is to learn better point-wise shape descriptors using human models
with ground truth correspondence. 
Several neural network based methods have recently been developed to this end to analyze meshes~\cite{Rodola14,MasBosBroVan15,Bosciani16,Monti17} or depth maps~\cite{wei2016dense}.
One can further improve these results by leveraging global context, for example, by estimating
an inter-surface functional map~\cite{Litany17}. These methods still rely on hand-crafted point-wise descriptors~\cite{Tombari10}
as input and use neural networks to improve results. The resulting functional maps only align basis functions and additional optimization
is required to extract consistent point-to-point correspondences~\cite{Ovsjanikov12}. One would also need 
to optimize for template deformation to use these matching techniques for surface reconstruction. 
In contrast our method does not rely on hand-crafted features (it only takes point coordinates as input) 
and implicitly learns a human body representation. It also directly outputs a template deformation.%

\subsubsection{Deep Learning for 3D data.} Following the success of deep learning approaches for image analysis, many techniques have been developed for processing 3D data, going beyond local descriptor learning to improve classification, segmentation, and reconstruction tasks. 
Existing networks operate on various shape representations, such as volumetric grids~\cite{Girdhar16b,wu20153d}, point clouds~\cite{qi2016pointnet,Qi:2017:nips,Fan:2017:cvpr}, geometry images \cite{Sinha2016,Sinha2017}, 
seamlessly parameterized surfaces~\cite{Maron17}, by aligning a shape to a grid via distance-preserving maps~\cite{Ezuz17}, by folding a surface~\cite{Yang2017FoldingNetPC} or by predicting chart representations~\cite{groueix2018}.
We build on these works in several ways. First, we process the point clouds representing the input shapes using an architecture similar to~\cite{qi2016pointnet}. Second, similar to \cite{Sinha2017}, we learn a surface representation. However, we do not explicitly encode correspondences in the output of a convolution network, but implicitly learn them by optimizing for parameters of the generation network as we optimize for reconstruction losss. 

\newcommand{\groundtruth}{\mathcal{S}}
\newcommand{\patchinput}{\mathcal{A}}
\newcommand{\shapefeature}{\mathbf{x}}
\newcommand{\mlp}{\mathcal{D}}
\newcommand{\encoder}{\mathcal{E}}
\newcommand{\paramsencoder}{\phi}
\newcommand{\lossmlp}{\mathcal{L}}
\newcommand{\lossatlas}{\mathcal{L}^\prime}
\newcommand{\parameters}{\theta}
\newcommand{\parameterset}{\Theta}
\newcommand{\atlaspoint}{\mathbf{p}}
\newcommand{\surfacepoint}{\mathbf{q}}
\newcommand{\reference}{r}
\newcommand{\target}{t}
\newcommand{\correspondences}{\mathcal{C}}
\newcommand{\losstrainsup}{\mathcal{L}^{\textrm{sup}}}
\newcommand{\losscham}{\mathcal{L}^{\textrm{CD}}}
\newcommand{\lossedges}{\mathcal{L}^{\textrm{edges}}}
\newcommand{\losslap}{\mathcal{L}^{\textrm{Lap}}}
\newcommand{\losstrainunsup}{\mathcal{L}^{\textrm{unsup}}}
\newcommand{\lossrecon}{\mathcal{L}^{\textrm{CD}}}

\newcommand{\regularizer}{\mathcal{R}}

\begin{figure}[!h]
\centering
\begin{subfigure}[b]{0.9\linewidth}
 \includegraphics[ width=\linewidth]{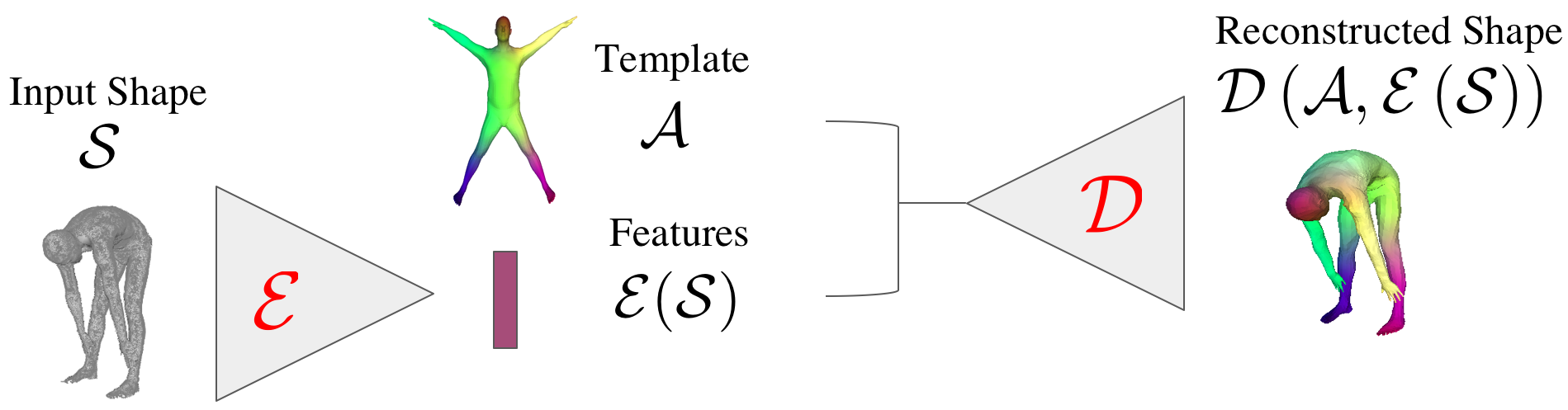}%
 \caption{\bf Network training}
\end{subfigure}\\
\begin{subfigure}[b]{0.85\linewidth}
 \includegraphics[ width=\linewidth]{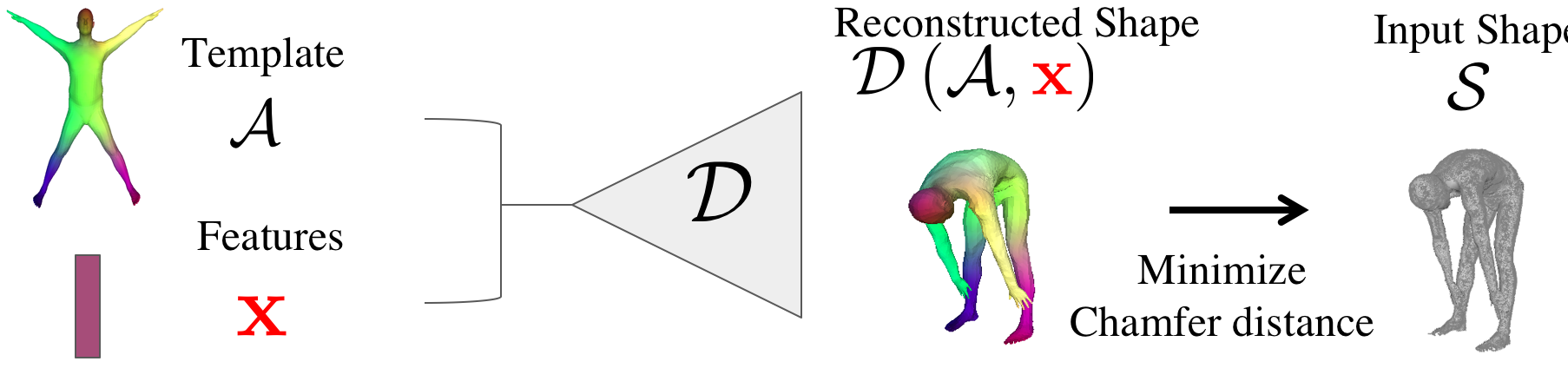}%
 \caption{\bf  Local optimization of feature {\color{red} $\shapefeature$}}
\end{subfigure}\\
\begin{subfigure}[b]{0.85\linewidth}
 \includegraphics[ width=\linewidth]{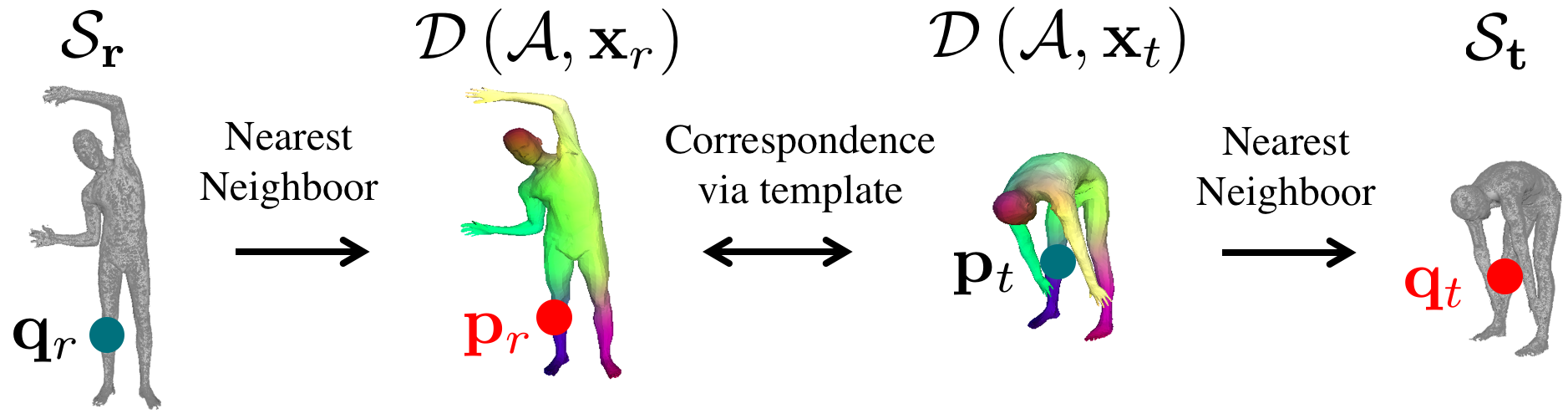}%
 \caption{\bf Correspondences}
\end{subfigure}
\caption{\textbf{Method overview.} (a) A feed-forward pass in our autoencoder encodes input point cloud $\mathbf{\groundtruth}$ to latent code $\encoder\left(\groundtruth\right)$ and reconstruct $\mathbf{\groundtruth}$ using $\encoder\left(\groundtruth\right)$ to deform the template $\mathbf{\patchinput}$. (b) We refine the reconstruction $\mlp\left(\mathbf{\patchinput},\encoder\left(\groundtruth\right)\right) $ by performing a regression step over the latent variable  $\shapefeature$, minimizing the Chamfer distance between $\mlp\left(\mathbf{\patchinput},\shapefeature\right)$ and $\mathbf{\groundtruth}$. (c) Finally, given two point clouds $\mathbf{\groundtruth_\reference}$ and $\mathbf{\groundtruth_\target}$, to match a point $\surfacepoint_\reference$ on $\mathbf{\groundtruth_\reference}$ to a point $\surfacepoint_\target$ on $\mathbf{\groundtruth_\target}$, we look for the nearest neighbor $ \atlaspoint_\reference$ of $\surfacepoint_\reference$ in $\mlp\left(\mathbf{\patchinput},\shapefeature_\reference\right)$, which is by design in correspondence with  $\atlaspoint_\target$; and look for the nearest neighbor $\surfacepoint_\target$ of $\atlaspoint_\target$ on   $\mathbf{\groundtruth_\target}$. {\color{red} Red} indicates what is being optimised.}
\label{fig:overview}
\end{figure}
\section{Method}

Our goal is, given a reference shape $\groundtruth_\reference$ and a target shape $\groundtruth_\target$, to return a set of point correspondences $\correspondences$ between the shapes. 
We do so using two key ideas. First, we learn to predict a transformation between the shapes instead of directly learning the correspondences. This transformation, from 3D to 3D can indeed be represented by a neural network more easily than the association between variable and large number of points. The second idea is to learn transformations only from one template $\patchinput$ to any shape.
Indeed, the large variety of possible poses of humans makes considering all pairs of possible poses intractable during training. %
We instead decouple the correspondence problem into finding two sets of correspondences to a common template shape. 
We can then form our final correspondences between the input shapes via indexing through the template shape. 
An added benefit is during training we simply need to vary the pose for a single shape and use the known correspondences to the template shape as the supervisory signal.

Our approach has three main steps which are visualized figure \ref{fig:overview}.  First, a feed-forward pass through our encoder network generates an initial global shape descriptor (Section~\ref{sec:reconstruction}). Second, we use gradient descent through our decoder Shape Deformation Network to refine this shape descriptor to improve the reconstruction quality (Section~\ref{sec:regression}). We can then use the template to match points between any two input shapes (Section~\ref{sec:correspondences}).

\subsection{Learning 3D shape reconstruction by template deformation}
\label{sec:reconstruction}

To put an input shape $\groundtruth$ in correspondence with a template $\patchinput$, our first goal is to design a neural network that will take $\groundtruth$ as input and predict transformation parameters. We do so by training an encoder-decoder architecture. The encoder  $\encoder_{\paramsencoder}$ defined by its parameters $\paramsencoder$ takes as input 3D points, and is a simplified version of the network presented in \cite{qi2016pointnet}. It applies to each input 3D point coordinate a multi-layer perceptron with hidden feature size of 64, 128 and 1024, then maxpooling over the resulting features over all points followed by a linear layer, leading to feature of size 1024 $\encoder_{\paramsencoder}\left(\groundtruth\right)$. This feature, together with the 3D coordinates of a point on the template $\atlaspoint\in\patchinput$, are taken as input to the decoder $\mlp_{\parameters}$ with parameters $\parameters$, which is trained to predict the position $\surfacepoint$ of the corresponding point in the input shape. This decoder Shape Deformation Network is a multi-layer perceptron with hidden layers of size 1024, 512, 254 and 128, followed by a hyperbolic tangent. This architecture maps any points from the template domain to the reconstructed surface. By sampling the template more or less densely, we can generate an arbitrary number of output points by sequentially applying the decoder over sampled template points.

This encoder-decoder architecture is trained end-to-end. We assume that  we are given as input a training set of $N$ shapes $\left\{\groundtruth^{\left(i\right)}\right\}_{i=1}^N$ with 
each shape having a set of $P$ vertices $\left\{\surfacepoint_j\right\}_{j=1}^P$. We consider two training scenarios: one where the correspondences between the template and the training shapes are known (supervised case) and one where they are unknown (unsupervised case). Supervision is typically available if the training shapes are generated by deforming a parametrized template, but real object scans are typically obtained without correspondences.

\subsubsection{Supervised loss.}
In the supervised case, we assume that for each point $\surfacepoint_j$ on a training shape we know the correspondence $\atlaspoint_j\leftrightarrow\surfacepoint_j$ to a point $\atlaspoint_j\in\patchinput$ on the template $\patchinput$. Given these training correspondences, we learn the encoder $\encoder_{\paramsencoder}$ and decoder $\mlp_{\parameters}$ by simply optimizing the following reconstruction losses,
\begin{equation}
\losstrainsup(\parameters,\paramsencoder) = \sum_{i=1}^N \sum_{j=1}^P | \mlp_\parameters\left(\atlaspoint_j; \encoder_{\paramsencoder}\left(\groundtruth^{\left(i\right)}\right)\right) - \surfacepoint^{\left(i\right)}_{j} |^2
\label{eqn:training_sup}
\end{equation}
where the sums are over all $P$ vertices of all $N$ example shapes.

\subsubsection{Unsupervised loss.}
In the case where correspondences between the exemplar shapes and the template are not available, we also optimize the reconstructions, but also regularize the deformations toward isometries. For reconstruction, we use the Chamfer distance $\losscham$ between the inputs $\groundtruth_i$ and reconstructed point clouds $\mlp_\parameters\left(\patchinput; \encoder_{\paramsencoder}\left(\groundtruth^{\left(i\right)}\right)\right)$. For regularization, we use two different terms. The first term $\losslap$ encourages the Laplacian operator defined on the template and the deformed template to be the same (which is the case for isometric deformations of the surface). The second term $\lossedges$ encourages the ratio between edges length in the template and its deformed version to be close to 1. More details on these different losses are given in supplementary material. The final loss we optimize is:
\begin{equation}
\losstrainunsup =  \losscham +\lambda_{Lap}\losslap +\lambda_{edges}\lossedges
\label{eqn:training_unsup}
\end{equation}
where $\lambda_{Lap}$ and $\lambda_{edges}$ control the influence of regularizations against the data term $\losscham$. They are both set to $5.10^{-3}$ in our experiments.\\

We optimize the loss using the Adam solver, with a learning rate of $10^{-3}$ for 25 epochs then $10^{-4}$ for 2 epochs, batches of 32 shapes, and 6890 points per shape. 

One interesting aspect of our approach is that it learns jointly a parameterization of the input shapes via the decoder and to perdict the parameters $\encoder_{\paramsencoder}\left(\groundtruth\right)$ for this parameterization via the encoder. However, the predicted parameters $\encoder_{\paramsencoder}\left(\groundtruth\right)$ for an input shape $\groundtruth$ are not necessarily optimal, because of the limited power of the encoder. Optimizing these parameters turns out to be important for the final results, and is the focus of the second step of our pipeline. 

\subsection{Optimizing shape reconstruction}
\label{sec:regression}

We now assume that we are given a shape $\groundtruth$ as well as learned weights for the encoder $\encoder_{\paramsencoder}$ and decoder $\mlp_{\parameters}$ networks. To find correspondences between the template shape and the input shape, we will use a nearest neighbor search to find correspondences between that input shape and its reconstruction.
For this step to work, we need the reconstruction to be accurate. The reconstruction given by the parameters $\encoder_{\paramsencoder}\left(\groundtruth\right)$ is only approximate and can be improved. Since we do not know correspondences between the input and the generated shape, we cannot minimize the loss given in equation (\ref{eqn:training_sup}), which requires correspondences. Instead, we minimize with respect to the global feature $\shapefeature$ the Chamfer distance between the reconstructed shape and the input:
\begin{equation}
    \lossrecon(\shapefeature; \groundtruth) = \sum_{\atlaspoint\in\patchinput}  \min_{\surfacepoint\in\groundtruth} \left|\mlp_{\parameters}\left(\atlaspoint; \shapefeature\right) - \surfacepoint\right|^2
    + \sum_{\surfacepoint\in\groundtruth}  \min_{\atlaspoint\in\patchinput}\left|\mlp_{\parameters}\left(\atlaspoint; \shapefeature\right) - \surfacepoint\right|^2.
\label{eqn:atlas_loss}
\end{equation}

Starting from the parameters predicted by our first step $\shapefeature = \encoder_{\paramsencoder}\left(\groundtruth\right)$, we optimize this loss using the Adam solver for 3,000 iterations with learning rate $5*10^{-4}$. %
Note that the good initialization given by our first step is key since Equation(~\ref{eqn:atlas_loss}) corresponds to a highly non-convex problem, as shown in Figure~\ref{fig:regression}.

\subsection{Finding 3D shape correspondences}
\label{sec:correspondences}

To recover correspondences between two 3D shapes $\groundtruth_\reference$ and $\groundtruth_\target$, we first compute the parameters to deform the template to these shapes, $\shapefeature_\reference$ and $\shapefeature_\target$, using the two steps outlined in section~\ref{sec:reconstruction} and~\ref{sec:regression}.
Next, given a 3D point $\surfacepoint_\reference$ on the reference shape $\groundtruth_\reference$, we first find the point $\atlaspoint$ on the template $\patchinput$ such that its transformation with parameters $\shapefeature_\reference$, $\mlp_{\parameters}\left(\atlaspoint; \shapefeature_\reference\right)$ is closest to $\surfacepoint_\reference$. 
Finally we find the 3D point $\surfacepoint_\target$ on the target shape $\groundtruth_\target$ that is the closest to the transformation of $\atlaspoint$ with parameters $\shapefeature_\target$, $\mlp_{\parameters}\left(\atlaspoint; \shapefeature_\target\right)$. %
 Our algorithm is summarized in Algorithm~\ref{alg:inference} and illustrated in Figure~\ref{fig:overview}.

\begin{algorithm}[t]
\caption{Algorithm for finding 3D shape correspondences}\label{alg:inference}
\SetKwInOut{Input}{Input}
\SetKwInOut{Output}{Output}
\Input{Reference shape $\groundtruth_\reference$ and target shape $\groundtruth_\target$}
\Output{Set of 3D point correspondences $\correspondences$}
\#Regression steps over latent code to find best reconstruction of $\groundtruth_\reference$ and $\groundtruth_\target$ \\
$\shapefeature_\reference\leftarrow\argmin_{\shapefeature} \lossrecon\left(\shapefeature; \groundtruth_\reference\right)$  \#detailed in equation~(\ref{eqn:atlas_loss})\\
$\shapefeature_\target\leftarrow\argmin_{\shapefeature} \lossrecon\left(\shapefeature; \groundtruth_\target\right)$ \#detailed in equation~(\ref{eqn:atlas_loss}) \\
$\correspondences\leftarrow\varnothing$ \\
\# Matching of $\surfacepoint_\reference\in\groundtruth_\reference$ to  $\surfacepoint_\target\in\groundtruth_\target$ \\
\ForEach{$\surfacepoint_\reference\in\groundtruth_\reference$}
{
    $\atlaspoint\leftarrow\argmin_{\atlaspoint^\prime\in\patchinput} |\mlp_{\parameters}\left(\atlaspoint^\prime; \shapefeature_\reference\right)-\surfacepoint_\reference|^2$ \\
    $\surfacepoint_\target\leftarrow\argmin_{\surfacepoint^\prime\in\groundtruth_\target} |\mlp_{\parameters}\left(\atlaspoint; \shapefeature_\target\right)-\surfacepoint^\prime|^2$ \\
    $\correspondences\leftarrow\correspondences\cup\left\{\left(\surfacepoint_\reference,\surfacepoint_\target\right)\right\}$
}
return $\correspondences$
\end{algorithm}

\noindent
\section{Results}

\subsection{Datasets}

\label{sec:data}

\begin{figure}[t]
\centering
\begin{subfigure}[b]{0.32\linewidth}%
\centering
 \includegraphics[height=50pt]{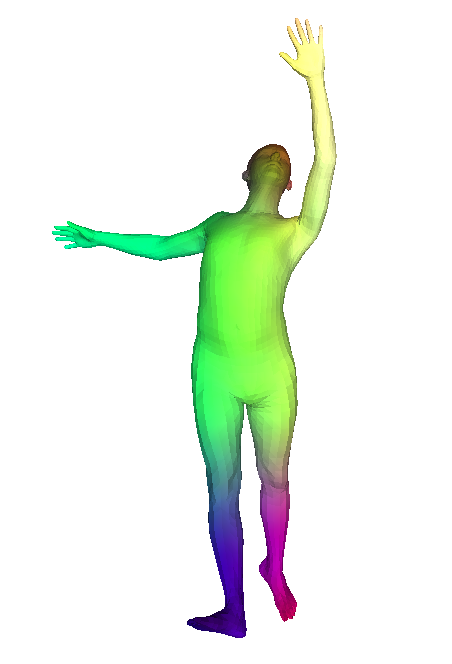}
  \includegraphics[height=50pt]{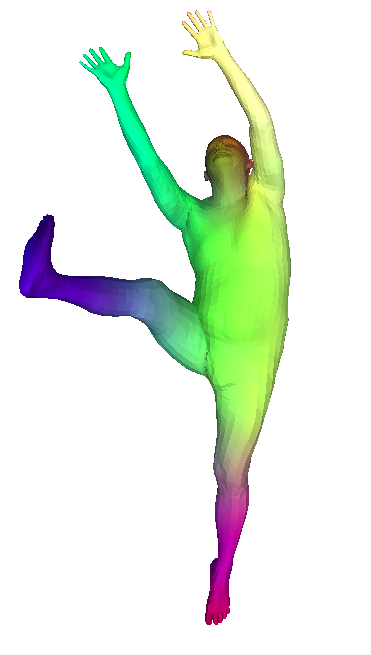}
 \includegraphics[height=50pt]{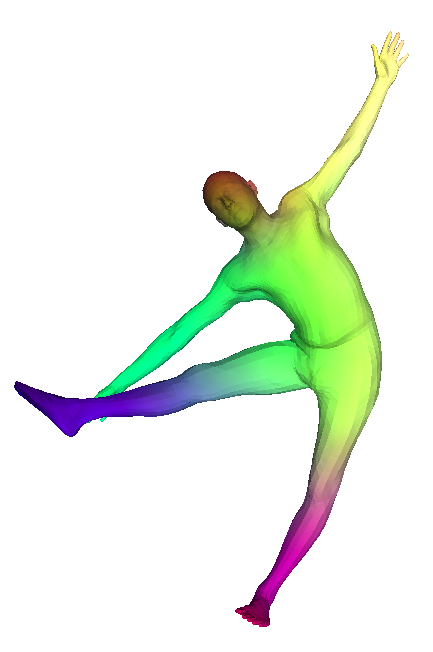}
 \caption{SURREAL~\cite{varol17a}}
\end{subfigure}
\vline
\begin{subfigure}[b]{0.32\linewidth}
\centering
 \includegraphics[height=45pt]{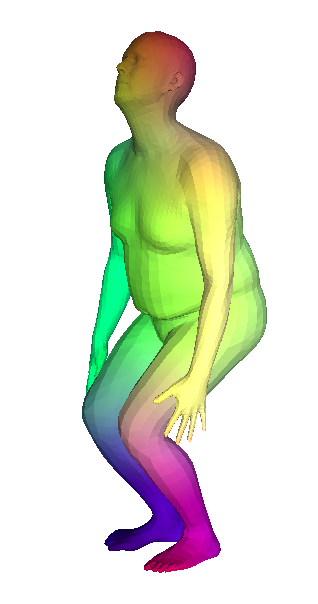}
  \includegraphics[height=45pt]{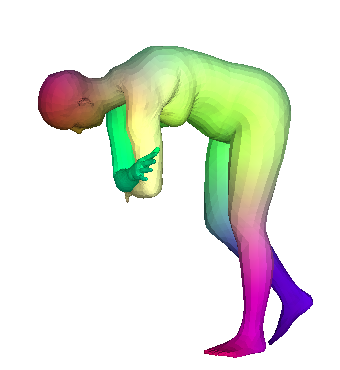}
 \includegraphics[height=45pt]{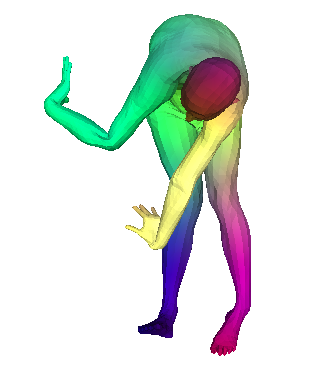}
  \caption{Bent shapes}
 \end{subfigure}
\vline
\begin{subfigure}[b]{0.32\linewidth}
\centering
 \includegraphics[height=45pt]{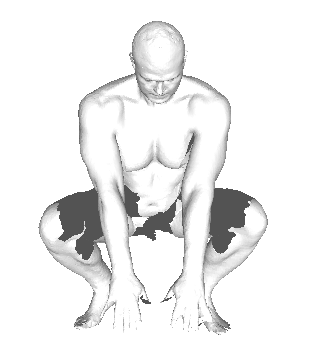}%
 \includegraphics[height=45pt]{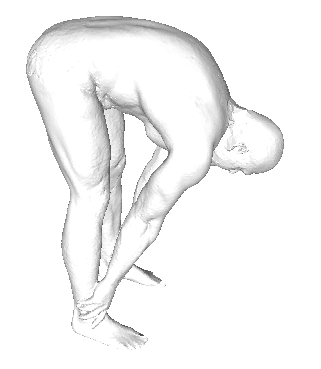}%
 \includegraphics[height=50pt]{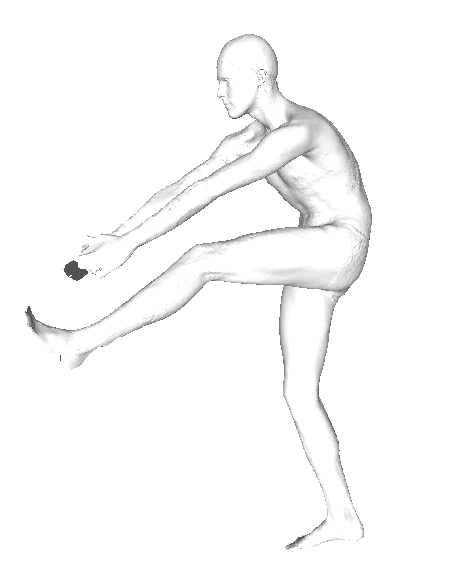}%
  \caption{FAUST~\cite{Bogo:CVPR:2014}}
\end{subfigure}

\caption{
{\bf Examples of the different datasets used in the paper.%
}}
  \label{fig:bent}
\end{figure}

\subsubsection{Synthetic training data.}
To train our algorithm, we require a large set of shapes. %
We thus rely on synthetic data for training our model. 

For human shapes, we use SMPL~\cite{Bogo:CVPR:2014}, a state-of-the-art generative model for synthetic humans. To obtain realistic human body shape and poses from the SMPL model, we sampled $2.10^5$ %
parameters estimated in the SURREAL dataset~\cite{varol17a}. %
One limitation of the SURREAL dataset is it does not include any humans bent over. 
Without adapted training data, our algorithm generalized poorly to these poses. To overcome this limitation, we generated an extension of the dataset. We first manually estimated 7 key-joint parameters (among 23 joints in the SMPL skeletons) to generate bent humans. We then sampled randomly the 7 parameters around these values, and used parameters from the SURREAL dataset for the other pose and body shape parameters. Note that not all meshes generated with this strategy are realistic as shown in  figure~\ref{fig:bent}. They however allow us to better cover the space of possible poses, and we added $3 \cdot 10^4$ shapes generated with this method to our dataset.
Our final dataset thus has $2.3 \cdot 10^5$ human meshes with a large variety of realistic poses and body shapes.%

For animal shapes, we use the %
SMAL \cite{Zuffi:CVPR:2017} model, which provides the equivalent of SMPL for several animals. Recent papers estimate model parameters from images, but no large-scale parameter set is yet available. For training we thus generated models from SMAL with random parameters (drawn from a Gaussian distribution of \textit{ad-hoc} variance 0.2). This approach works for the 5 categories available in SMAL. In SMALR~\cite{Zuffi:CVPR:2018}, Zuffi et al. showed that the SMAL model could be generalized to other animals using only an image dataset as input, demonstrating it on 17 additional categories. Note that since the templates for two animals are in correspondences, our method can be used to get inter-category correspondences for animals. We qualitatively demonstrate this on hippopotamus/horses in the appendix~\cite{appendix}.

\subsubsection{Testing data.}
We evaluate our algorithm on the FAUST~\cite{Bogo:CVPR:2014}, TOSCA~\cite{bronstein2008numerical} and SCAPE~\cite{anguelov2005scape} datasets. 

The FAUST dataset consists of 100 training and 200 testing scans of approximately 170,000 vertices. They may include noise and have holes, typically missing part of the feet. In this paper, we never used the training set, except for a single baseline experiment, and we focus on the test set. %
Two challenges are available, focusing on intra- and inter-subject correspondences. %
The error is the average Euclidean distance between the estimated projection and the ground-truth projection. We evaluated our method through the online server and are the best public results on the 'inter' challenge at the time of submission\footnote{\url{http://faust.is.tue.mpg.de/challenge/Inter-subject\_challenge}}.

The SCAPE~\cite{anguelov2005scape} dataset has two sets of 71 meshes : the first set consists of real scans with holes and occlusions and the second set are registered meshes aligned to the first set. %
The poses are  different from both our training dataset and FAUST.

TOSCA is a dataset produced by deforming 3 template meshes (human, dog, and horse). Each mesh is deformed into multiple poses, and might have various additional perturbations such as random holes in the surface, local and global scale variations, noise in vertex positions, varying sampling density, and changes in topology. %

\subsubsection{Shape normalization.} To be processed and reconstructed by our network, the training and testing shapes must be normalized in a similar way. Since the vertical direction is usually known, we used synthetic shapes with approximately the same vertical axis. We also kept a fixed orientation around this vertical axis, and at test time selected the one out of 50 different orientations which leads to the smaller reconstruction error in term of Chamfer distance. %
Finally, we centered all meshes according to the center of their bounding box and, for the training data only, added a random translation in each direction sampled uniformly between -3cm and 3cm to increase robustness.

\subsection{Experiments}

In this part, we %
 analyze the key components of our pipeline. %
 More results are available in the appendix~\cite{appendix}.

\subsubsection{Results on FAUST.} The method presented above leads to the best results to date on the FAUST-inter dataset:  2.878 cm : {\bf an improvement of 8\% over state of the art}, 3.12cm for \cite{Zuffi15} and 4.82cm for \cite{Litany17}. 
Although it cannot take advantage of the fact that two meshes represent the same person%
, our method is also the second best performing (average error of 1.99 cm) on FAUST-intra challenge.

\begin{figure}[th]
\centering
\begin{subfigure}[b]{0.28\linewidth}
\centering
 \includegraphics[width=0.45\linewidth]{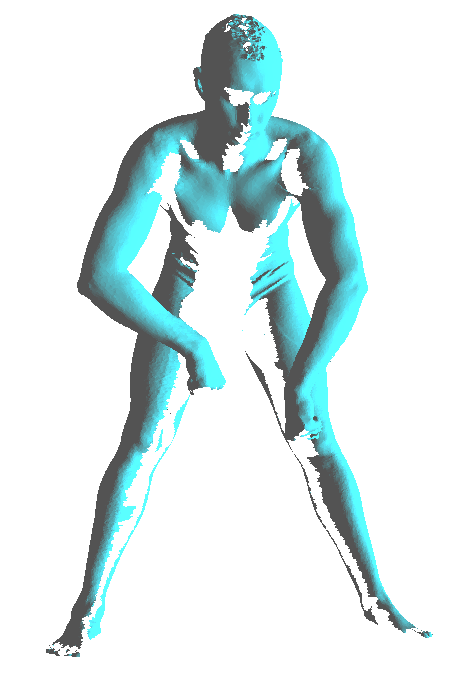}
 \includegraphics[width=0.45\linewidth]{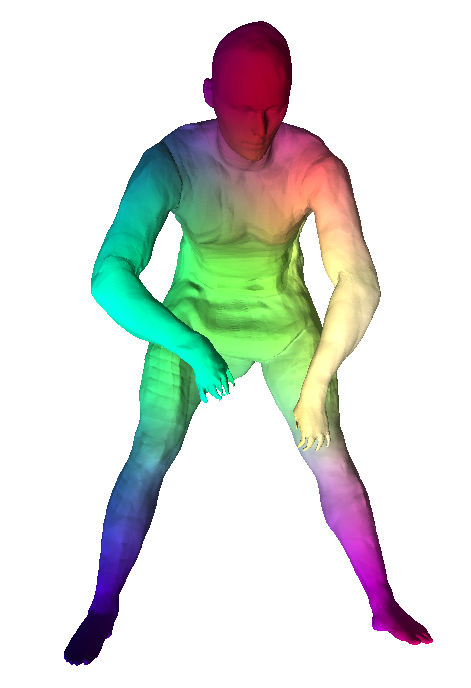}\\
 \includegraphics[width=0.45\linewidth]{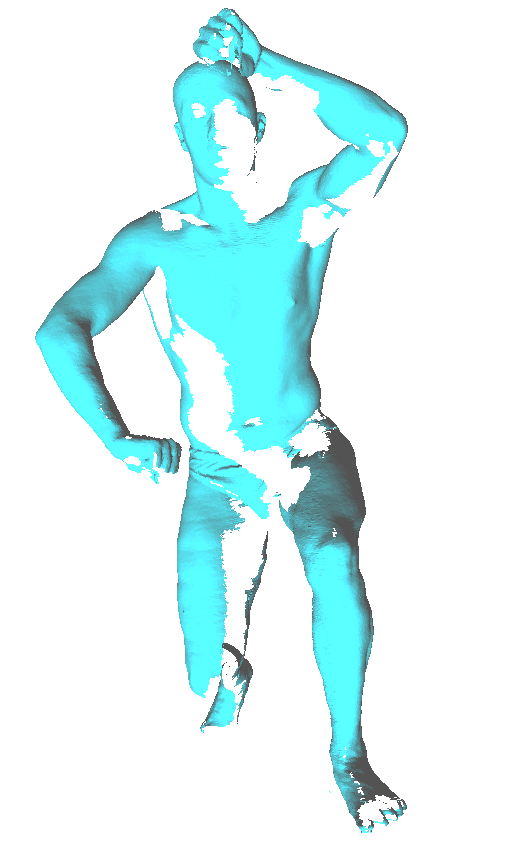}
  \includegraphics[width=0.45\linewidth]{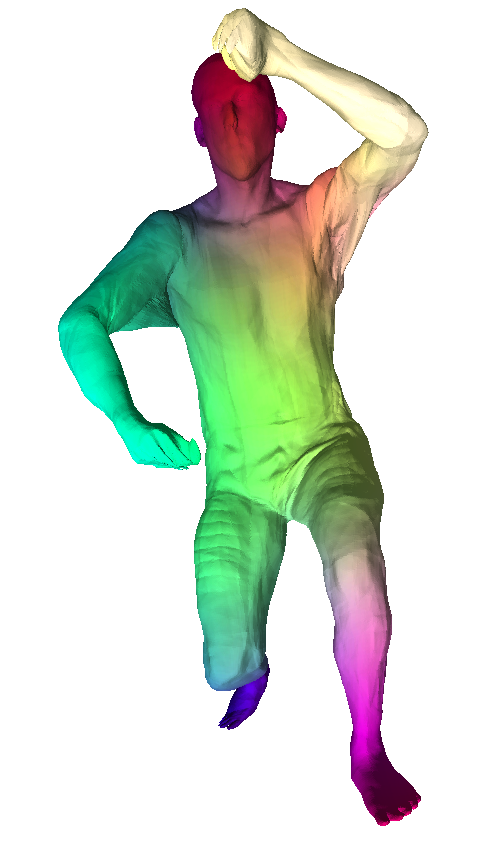}
 \caption{SCAPE~\cite{anguelov2005scape}\label{fig:SCAPE}}
\end{subfigure}~~~~
\begin{subfigure}[b]{0.255\linewidth}
\centering
 \includegraphics[width=0.45\linewidth]{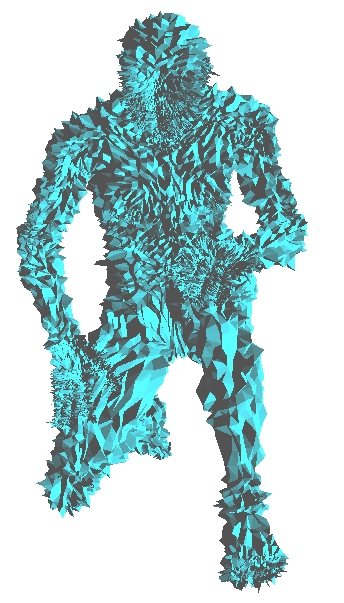}
 \includegraphics[width=0.45\linewidth]{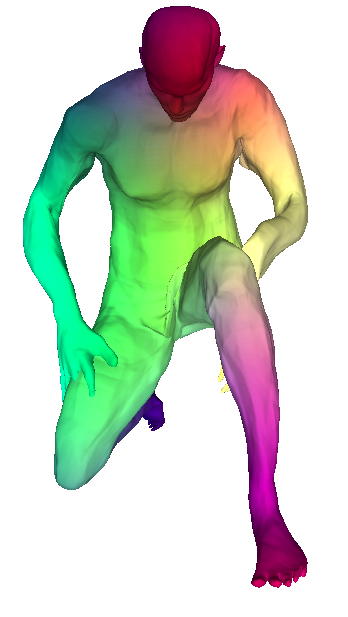}\\
 \includegraphics[width=0.45\linewidth]{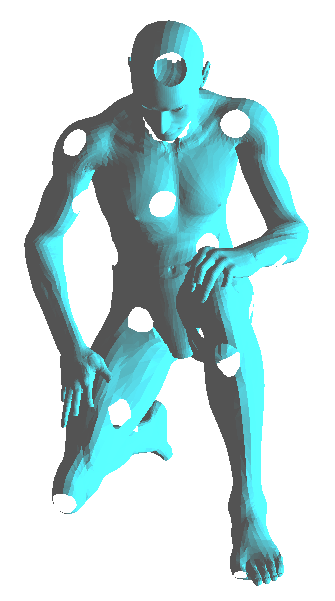}
 \includegraphics[width=0.45\linewidth]{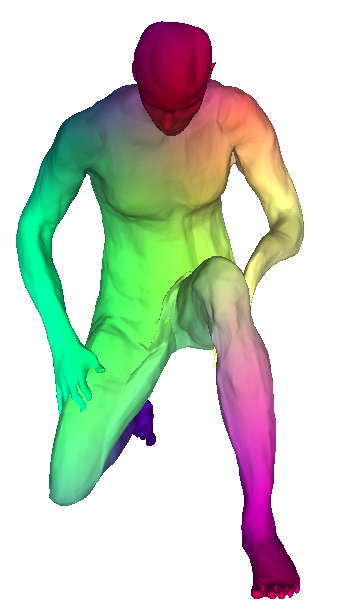}
  \caption{TOSCA~\cite{bronstein2008numerical}\label{fig:TOSCA}}
 \end{subfigure}~~
\begin{subfigure}[b]{0.36\linewidth}
\centering
 \includegraphics[width=0.38\linewidth]{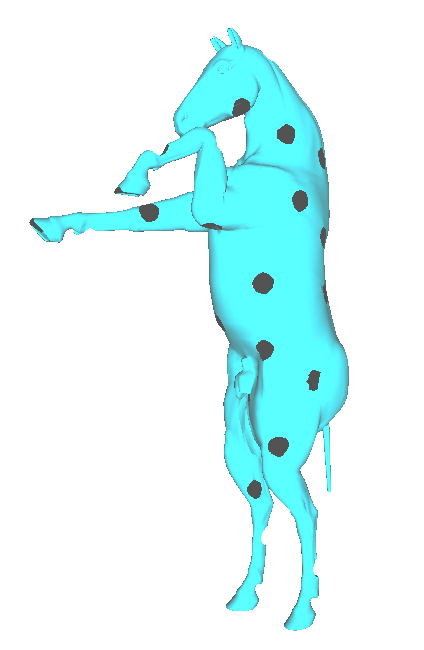}
 \includegraphics[width=0.38\linewidth]{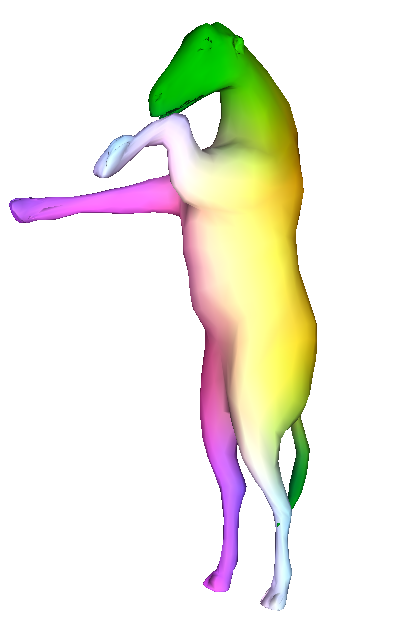}\\
 \includegraphics[width=0.45\linewidth]{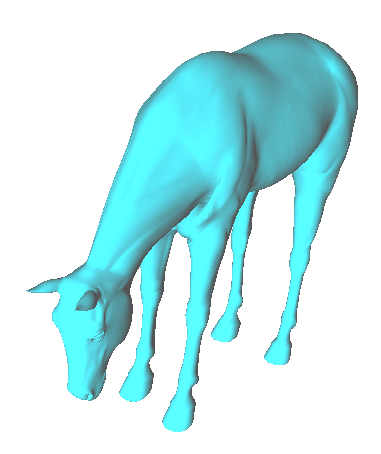}
 \includegraphics[width=0.45\linewidth]{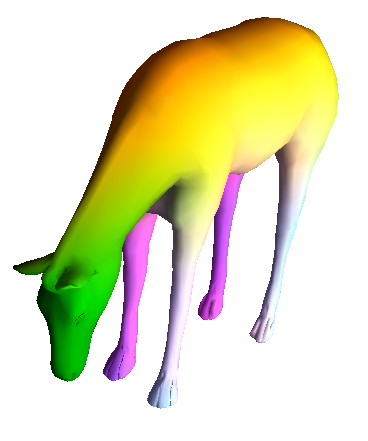}
  \caption{TOSCA animals~\cite{bronstein2008numerical}\label{fig:Animals}}
\end{subfigure}
\caption{ {\bf Other datasets.} Left images show the input, right images the reconstruction with colors showing correspondences. Our method works with real incomplete scans \textbf{(a)}, strong synthetic perturbations \textbf{(b)}, and on non-human shapes \textbf{(c)}.}
  \label{fig:rebuttal}
  
\end{figure}

\begin{figure}[th]
  \centering
\definecolor{mycolor1}{rgb}{0.60000,0.29804,0.00000}%
\definecolor{mycolor2}{rgb}{0.49804,0.00000,1.00000}%
\definecolor{mycolor3}{rgb}{1.00000,0.84314,0.00000}%
\begin{tikzpicture}

\tikzstyle{every node}=[font=\scriptsize]

\begin{axis}[%
width=0.5\linewidth,%
height=0.3\linewidth,%
scale only axis,
xmin=0,
xmax=0.1,
xlabel style={at={(0.5,0.02)}},
xlabel={\normalsize Geodesic error},
every x tick label/.append style={font=\color{black}, font=\footnotesize},
every y tick label/.append style={font=\color{black}, font=\footnotesize},
xmajorgrids,
xtick={0,0.02,0.04,0.06,0.08,0.1},
xticklabels = {0,0.02,0.04,0.06,0.08,0.1},
yticklabels = {0,0,20,40,60,80,100},
ymin=0,
ymax=1,
ylabel style={at={(0.04,0.5)}},
ylabel={\normalsize \% Correspondences},
ymajorgrids,
axis background/.style={fill=white},
legend style={
	at={(0.99,0.01)},
	anchor=south east,
	legend cell align=left,
	align=left
}
]
\addplot [color=red,solid,line width=2.0pt]
  table[row sep=crcr]{%
0.0   0.1008266667\\
0.001   0.1008266667\\
0.002   0.1008266667\\
0.003   0.1012355556\\
0.004   0.1023466667\\
0.005   0.1051733333\\
0.006   0.1107111111\\
0.007   0.1203733333\\
0.008   0.1336000000\\
0.009   0.1527200000\\
0.01   0.1754222222\\
0.011   0.2026933333\\
0.012   0.2352266667\\
0.013   0.2675111111\\
0.014   0.3004177778\\
0.015   0.3317688889\\
0.016   0.3645955556\\
0.017   0.3953155556\\
0.018   0.4266400000\\
0.019   0.4560977778\\
0.02   0.4853955556\\
0.021   0.5127377778\\
0.022   0.5397333333\\
0.023   0.5655288889\\
0.024   0.5895466667\\
0.025   0.6133866667\\
0.026   0.6357866667\\
0.027   0.6559466667\\
0.028   0.6755022222\\
0.029   0.6941066667\\
0.03   0.7103111111\\
0.031   0.7253422222\\
0.032   0.7417688889\\
0.033   0.7559288889\\
0.034   0.7692533333\\
0.035   0.7816533333\\
0.036   0.7936266667\\
0.037   0.8046666667\\
0.038   0.8149066667\\
0.039   0.8247822222\\
0.04   0.8347200000\\
0.041   0.8438844444\\
0.042   0.8527555556\\
0.043   0.8604444444\\
0.044   0.8677066667\\
0.045   0.8749422222\\
0.046   0.8808444444\\
0.047   0.8865600000\\
0.048   0.8920266667\\
0.049   0.8973688889\\
0.05   0.9025511111\\
0.051   0.9070400000\\
0.052   0.9116177778\\
0.053   0.9156800000\\
0.054   0.9196888889\\
0.055   0.9232977778\\
0.056   0.9267644444\\
0.057   0.9298311111\\
0.058   0.9328711111\\
0.059   0.9359644444\\
0.06   0.9386755556\\
0.061   0.9413244444\\
0.062   0.9438222222\\
0.063   0.9459644444\\
0.064   0.9479466667\\
0.065   0.9498666667\\
0.066   0.9519733333\\
0.067   0.9537066667\\
0.068   0.9555022222\\
0.069   0.9569955556\\
0.07   0.9588000000\\
0.071   0.9604444444\\
0.072   0.9620266667\\
0.073   0.9632888889\\
0.074   0.9645155556\\
0.075   0.9657777778\\
0.076   0.9669955556\\
0.077   0.9682400000\\
0.078   0.9695555556\\
0.079   0.9707288889\\
0.08   0.9719111111\\
0.081   0.9729066667\\
0.082   0.9740533333\\
0.083   0.9751377778\\
0.084   0.9760088889\\
0.085   0.9768000000\\
0.086   0.9775644444\\
0.087   0.9783022222\\
0.088   0.9790133333\\
0.089   0.9797600000\\
0.09   0.9806222222\\
0.091   0.9812266667\\
0.092   0.9819288889\\
0.093   0.9825777778\\
0.094   0.9831733333\\
0.095   0.9836800000\\
0.096   0.9842044444\\
0.097   0.9848000000\\
0.098   0.9853866667\\
0.099   0.9858666667\\
0.1   0.9863200000\\
};
\addlegendentry{Ours}

\addplot [color=black,solid,line width=2.0pt]
  table[row sep=crcr]{%
0.0	0.131128888888889\\
0.001	0.131128888888889\\
0.002	0.133084444444444\\
0.003	0.145804444444444\\
0.004	0.181511111111111\\
0.005	0.241235555555556\\
0.006	0.313635555555556\\
0.007	0.379395555555556\\
0.008	0.437413333333333\\
0.009	0.486746666666667\\
0.01	0.531271111111111\\
0.011	0.572488888888889\\
0.012	0.608106666666667\\
0.013	0.640266666666667\\
0.014	0.668506666666667\\
0.015	0.693208888888889\\
0.016	0.714568888888889\\
0.017	0.732986666666667\\
0.018	0.749475555555556\\
0.019	0.764995555555556\\
0.02	0.778737777777778\\
0.021	0.791431111111111\\
0.022	0.803555555555556\\
0.023	0.813413333333333\\
0.024	0.822968888888889\\
0.025	0.831413333333333\\
0.026	0.838977777777778\\
0.027	0.846213333333333\\
0.028	0.853217777777778\\
0.029	0.859653333333333\\
0.03	0.865866666666667\\
0.031	0.871662222222222\\
0.032	0.877075555555556\\
0.033	0.882417777777778\\
0.034	0.887333333333333\\
0.035	0.891733333333333\\
0.036	0.89616\\
0.037	0.900533333333333\\
0.038	0.904506666666667\\
0.039	0.908577777777778\\
0.04	0.912817777777778\\
0.041	0.916604444444444\\
0.042	0.920684444444444\\
0.043	0.92424\\
0.044	0.927413333333333\\
0.045	0.930311111111111\\
0.046	0.933048888888889\\
0.047	0.935955555555556\\
0.048	0.938657777777778\\
0.049	0.941208888888889\\
0.05	0.943893333333333\\
0.051	0.94616\\
0.052	0.948266666666667\\
0.053	0.950524444444445\\
0.054	0.952684444444445\\
0.055	0.954711111111111\\
0.056	0.956613333333333\\
0.057	0.958497777777778\\
0.058	0.959991111111111\\
0.059	0.961475555555556\\
0.06	0.962648888888889\\
0.061	0.963884444444444\\
0.062	0.965093333333333\\
0.063	0.96608\\
0.064	0.967004444444444\\
0.065	0.967795555555556\\
0.066	0.96856\\
0.067	0.969244444444444\\
0.068	0.969644444444444\\
0.069	0.970088888888889\\
0.07	0.970346666666667\\
0.071	0.970568888888889\\
0.072	0.970817777777778\\
0.073	0.971013333333333\\
0.074	0.971235555555556\\
0.075	0.971368888888889\\
0.076	0.971528888888889\\
0.077	0.971617777777778\\
0.078	0.97176\\
0.079	0.971848888888889\\
0.08	0.971937777777778\\
0.081	0.972088888888889\\
0.082	0.972195555555556\\
0.083	0.972302222222222\\
0.084	0.972417777777778\\
0.085	0.972497777777778\\
0.086	0.972568888888889\\
0.087	0.972657777777778\\
0.088	0.9728\\
0.089	0.972915555555555\\
0.09	0.97304\\
0.091	0.973226666666666\\
0.092	0.973297777777778\\
0.093	0.973422222222222\\
0.094	0.973537777777778\\
0.095	0.973644444444444\\
0.096	0.973688888888889\\
0.097	0.973777777777778\\
0.098	0.973866666666667\\
0.099	0.973955555555555\\
0.1	0.97408\\
0.101	0.974195555555555\\
0.102	0.974328888888889\\
0.103	0.974506666666667\\
0.104	0.974675555555555\\
0.105	0.974826666666667\\
0.106	0.975048888888889\\
0.107	0.975173333333333\\
0.108	0.975262222222222\\
0.109	0.975413333333333\\
0.11	0.975582222222222\\
0.111	0.975733333333333\\
0.112	0.97592\\
0.113	0.976035555555555\\
0.114	0.97624\\
0.115	0.976346666666667\\
0.116	0.976435555555556\\
0.117	0.976542222222222\\
0.118	0.976657777777778\\
0.119	0.976728888888889\\
0.12	0.976844444444444\\
0.121	0.977004444444444\\
0.122	0.977137777777778\\
0.123	0.977262222222222\\
0.124	0.977333333333333\\
0.125	0.977422222222222\\
0.126	0.977502222222222\\
0.127	0.977608888888889\\
0.128	0.977688888888889\\
0.129	0.977751111111111\\
0.13	0.977813333333333\\
0.131	0.977893333333333\\
0.132	0.977955555555555\\
0.133	0.978053333333333\\
0.134	0.978142222222222\\
0.135	0.978213333333333\\
0.136	0.978257777777778\\
0.137	0.978266666666667\\
0.138	0.978275555555555\\
0.139	0.978284444444444\\
0.14	0.978284444444444\\
0.141	0.978284444444444\\
0.142	0.978284444444444\\
0.143	0.978284444444444\\
0.144	0.978284444444444\\
0.145	0.978284444444444\\
0.146	0.978284444444444\\
0.147	0.978284444444444\\
0.148	0.978284444444444\\
0.149	0.978284444444444\\
0.15	0.978284444444444\\
0.151	0.978284444444444\\
0.152	0.978284444444444\\
0.153	0.978284444444444\\
0.154	0.978284444444444\\
0.155	0.978293333333333\\
0.156	0.978302222222222\\
0.157	0.978302222222222\\
0.158	0.978302222222222\\
0.159	0.978302222222222\\
0.16	0.978311111111111\\
0.161	0.97832\\
0.162	0.978328888888889\\
0.163	0.978337777777778\\
0.164	0.978337777777778\\
0.165	0.978337777777778\\
0.166	0.978346666666667\\
0.167	0.978346666666667\\
0.168	0.978364444444444\\
0.169	0.978373333333333\\
0.17	0.978382222222222\\
0.171	0.978382222222222\\
0.172	0.978382222222222\\
0.173	0.978391111111111\\
0.174	0.9784\\
0.175	0.978408888888889\\
0.176	0.978435555555556\\
0.177	0.978462222222222\\
0.178	0.978488888888889\\
0.179	0.978542222222222\\
0.18	0.978595555555556\\
0.181	0.978613333333333\\
0.182	0.97864\\
0.183	0.978675555555556\\
0.184	0.978693333333333\\
0.185	0.97872\\
0.186	0.978737777777778\\
0.187	0.978764444444444\\
0.188	0.978782222222222\\
0.189	0.978782222222222\\
0.19	0.978808888888889\\
0.191	0.978826666666667\\
0.192	0.978862222222222\\
0.193	0.978871111111111\\
0.194	0.978897777777778\\
0.195	0.978897777777778\\
0.196	0.978906666666667\\
0.197	0.978924444444444\\
0.198	0.978942222222222\\
0.199	0.978942222222222\\
0.2	0.978942222222222\\
0.201	0.978942222222222\\
0.202	0.978951111111111\\
0.203	0.978951111111111\\
0.204	0.978951111111111\\
0.205	0.978951111111111\\
0.206	0.978951111111111\\
0.207	0.978951111111111\\
0.208	0.978951111111111\\
0.209	0.978951111111111\\
0.21	0.978951111111111\\
0.211	0.978951111111111\\
0.212	0.978951111111111\\
0.213	0.978951111111111\\
0.214	0.978951111111111\\
0.215	0.97896\\
0.216	0.978968888888889\\
0.217	0.978977777777778\\
0.218	0.978977777777778\\
0.219	0.978986666666667\\
0.22	0.978986666666667\\
0.221	0.978995555555556\\
0.222	0.978995555555556\\
0.223	0.978995555555556\\
0.224	0.978995555555556\\
0.225	0.978995555555556\\
0.226	0.978995555555556\\
0.227	0.978995555555556\\
0.228	0.978995555555556\\
0.229	0.978995555555556\\
0.23	0.978995555555556\\
0.231	0.978995555555556\\
0.232	0.978995555555556\\
0.233	0.978995555555556\\
0.234	0.978995555555556\\
0.235	0.978995555555556\\
0.236	0.978995555555556\\
0.237	0.978995555555556\\
0.238	0.978995555555556\\
0.239	0.978995555555556\\
0.24	0.978995555555556\\
0.241	0.978995555555556\\
0.242	0.978995555555556\\
0.243	0.978995555555556\\
0.244	0.978995555555556\\
0.245	0.978995555555556\\
0.246	0.978995555555556\\
0.247	0.978995555555556\\
0.248	0.978995555555556\\
0.249	0.978995555555556\\
0.25	0.978995555555556\\
0.251	0.979004444444444\\
0.252	0.979004444444444\\
0.253	0.979013333333333\\
0.254	0.979022222222222\\
0.255	0.979022222222222\\
0.256	0.97904\\
0.257	0.979057777777778\\
0.258	0.979057777777778\\
0.259	0.979066666666667\\
0.26	0.979066666666667\\
0.261	0.979075555555556\\
0.262	0.979093333333333\\
0.263	0.979111111111111\\
0.264	0.97912\\
0.265	0.979128888888889\\
0.266	0.979128888888889\\
0.267	0.979137777777778\\
0.268	0.979137777777778\\
0.269	0.979137777777778\\
0.27	0.979146666666667\\
0.271	0.979146666666667\\
0.272	0.979146666666667\\
0.273	0.979146666666667\\
0.274	0.979146666666667\\
0.275	0.979146666666667\\
0.276	0.979146666666667\\
0.277	0.979146666666667\\
0.278	0.979146666666667\\
0.279	0.979146666666667\\
0.28	0.979146666666667\\
0.281	0.979146666666667\\
0.282	0.979146666666667\\
0.283	0.979146666666667\\
0.284	0.979146666666667\\
0.285	0.979146666666667\\
0.286	0.979146666666667\\
0.287	0.979146666666667\\
0.288	0.979146666666667\\
0.289	0.979146666666667\\
0.29	0.979146666666667\\
0.291	0.979146666666667\\
0.292	0.979146666666667\\
0.293	0.979146666666667\\
0.294	0.979146666666667\\
0.295	0.979146666666667\\
0.296	0.979146666666667\\
0.297	0.979146666666667\\
0.298	0.979146666666667\\
0.299	0.979146666666667\\
0.3	0.979146666666667\\
0.301	0.979146666666667\\
};
\addlegendentry{FMNet~\cite{Litany17}}

\addplot [color=black!40!green,solid,line width=1.25pt]
  table[row sep=crcr]{%
0	0.0329955555555556\\
0.003	0.0337244444444444\\
0.006	0.0460622222222222\\
0.009	0.0856444444444444\\
0.012	0.135013333333333\\
0.015	0.179182222222222\\
0.018	0.223208888888889\\
0.021	0.268115555555556\\
0.024	0.311946666666667\\
0.027	0.352275555555556\\
0.03	0.388426666666667\\
0.033	0.422231111111111\\
0.036	0.453742222222222\\
0.039	0.482551111111111\\
0.042	0.509431111111111\\
0.045	0.534986666666667\\
0.048	0.558124444444444\\
0.051	0.579848888888889\\
0.054	0.59968\\
0.057	0.619164444444445\\
0.06	0.63776\\
0.063	0.655928888888889\\
0.066	0.672328888888889\\
0.069	0.688675555555555\\
0.072	0.70352\\
0.075	0.718177777777778\\
0.078	0.732897777777778\\
0.081	0.747386666666667\\
0.084	0.76048\\
0.087	0.773164444444445\\
0.09	0.784684444444444\\
0.093	0.79512\\
0.096	0.805324444444444\\
0.099	0.815128888888889\\
0.102	0.824506666666667\\
0.105	0.833173333333333\\
0.108	0.841173333333333\\
0.111	0.848168888888889\\
0.114	0.854942222222222\\
0.117	0.86144\\
0.12	0.867804444444444\\
0.123	0.874231111111111\\
0.126	0.879831111111111\\
0.129	0.885111111111111\\
0.132	0.890142222222222\\
0.135	0.89496\\
0.138	0.899653333333333\\
0.141	0.9036\\
0.144	0.907093333333333\\
0.147	0.910035555555555\\
0.15	0.913191111111111\\
0.153	0.916346666666667\\
0.156	0.919262222222222\\
0.159	0.922195555555555\\
0.162	0.924782222222223\\
0.165	0.92768\\
0.168	0.930231111111111\\
0.171	0.93272\\
0.174	0.93496\\
0.177	0.93736\\
0.18	0.939715555555556\\
0.183	0.942044444444444\\
0.186	0.944204444444444\\
0.189	0.946453333333333\\
0.192	0.948533333333333\\
0.195	0.950782222222222\\
0.198	0.952764444444445\\
0.201	0.954568888888889\\
0.204	0.956533333333333\\
0.207	0.958142222222222\\
0.21	0.960017777777778\\
0.213	0.961902222222222\\
0.216	0.963493333333333\\
0.219	0.965271111111111\\
0.222	0.966737777777778\\
0.225	0.968044444444444\\
0.228	0.969582222222222\\
0.231	0.971164444444444\\
0.234	0.972533333333333\\
0.237	0.973822222222222\\
0.24	0.974986666666667\\
0.243	0.976275555555556\\
0.246	0.977555555555556\\
0.249	0.978524444444445\\
0.252	0.979546666666667\\
0.255	0.980311111111111\\
0.258	0.981084444444444\\
0.261	0.981902222222222\\
0.264	0.982906666666667\\
0.267	0.983644444444444\\
0.27	0.9844\\
0.273	0.985084444444444\\
0.276	0.985768888888889\\
0.279	0.986444444444444\\
0.282	0.987004444444444\\
0.285	0.9876\\
0.288	0.988195555555556\\
0.291	0.988782222222222\\
0.294	0.989253333333333\\
0.297	0.989697777777778\\
0.3	0.990115555555556\\
};
\addlegendentry{GCNN \cite{masci15}}; %

\addplot [color=mycolor3,solid,line width=1.25pt]
  table[row sep=crcr]{%
0	0.0359733333333333\\
0.003	0.0362933333333333\\
0.006	0.0485066666666667\\
0.009	0.0995733333333333\\
0.012	0.1616\\
0.015	0.21528\\
0.018	0.266071111111111\\
0.021	0.319297777777778\\
0.024	0.367768888888889\\
0.027	0.41144\\
0.03	0.451173333333333\\
0.033	0.486106666666667\\
0.036	0.517475555555556\\
0.039	0.545582222222222\\
0.042	0.572257777777778\\
0.045	0.596257777777778\\
0.048	0.618497777777778\\
0.051	0.640195555555556\\
0.054	0.659137777777778\\
0.057	0.677742222222222\\
0.06	0.695146666666667\\
0.063	0.711822222222222\\
0.066	0.727848888888889\\
0.069	0.743431111111111\\
0.072	0.757528888888889\\
0.075	0.771244444444445\\
0.078	0.784248888888889\\
0.081	0.796542222222222\\
0.084	0.808266666666667\\
0.087	0.818728888888889\\
0.09	0.829475555555556\\
0.093	0.838471111111111\\
0.096	0.846506666666667\\
0.099	0.854746666666667\\
0.102	0.862062222222222\\
0.105	0.868515555555555\\
0.108	0.87472\\
0.111	0.880488888888889\\
0.114	0.885795555555556\\
0.117	0.890435555555556\\
0.12	0.895431111111111\\
0.123	0.900124444444444\\
0.126	0.904382222222222\\
0.129	0.908524444444444\\
0.132	0.912355555555556\\
0.135	0.915866666666667\\
0.138	0.918808888888889\\
0.141	0.921822222222222\\
0.144	0.924355555555555\\
0.147	0.926675555555555\\
0.15	0.928888888888889\\
0.153	0.931084444444444\\
0.156	0.933164444444444\\
0.159	0.935244444444445\\
0.162	0.937217777777778\\
0.165	0.939244444444445\\
0.168	0.940684444444444\\
0.171	0.942577777777778\\
0.174	0.944186666666667\\
0.177	0.945911111111111\\
0.18	0.947377777777778\\
0.183	0.94928\\
0.186	0.950764444444445\\
0.189	0.952346666666667\\
0.192	0.954177777777778\\
0.195	0.955768888888889\\
0.198	0.957146666666667\\
0.201	0.958675555555555\\
0.204	0.960168888888889\\
0.207	0.961493333333333\\
0.21	0.962666666666667\\
0.213	0.964133333333333\\
0.216	0.965528888888889\\
0.219	0.966844444444445\\
0.222	0.968177777777778\\
0.225	0.969653333333333\\
0.228	0.970684444444444\\
0.231	0.971955555555556\\
0.234	0.97304\\
0.237	0.974328888888889\\
0.24	0.975466666666667\\
0.243	0.976506666666667\\
0.246	0.977564444444444\\
0.249	0.978773333333333\\
0.252	0.979831111111111\\
0.255	0.980791111111111\\
0.258	0.981902222222222\\
0.261	0.983066666666667\\
0.264	0.984053333333333\\
0.267	0.985217777777778\\
0.27	0.986097777777778\\
0.273	0.986933333333333\\
0.276	0.987875555555556\\
0.279	0.988728888888889\\
0.282	0.989653333333334\\
0.285	0.990506666666667\\
0.288	0.991288888888889\\
0.291	0.992346666666667\\
0.294	0.993075555555556\\
0.297	0.993964444444444\\
0.3	0.994595555555555\\
};
\addlegendentry{LSCNN \cite{WFT2015}}; %

\addplot [color=black!20!blue,solid,line width=1.25pt]
  table[row sep=crcr]{%
0	0.0297155555555556\\
0.003	0.0301955555555556\\
0.006	0.0427377777777778\\
0.009	0.0857511111111111\\
0.012	0.137884444444444\\
0.015	0.184222222222222\\
0.018	0.230968888888889\\
0.021	0.278915555555556\\
0.024	0.323084444444444\\
0.027	0.363662222222222\\
0.03	0.401342222222222\\
0.033	0.436044444444444\\
0.036	0.467893333333333\\
0.039	0.496675555555556\\
0.042	0.525164444444445\\
0.045	0.5508\\
0.048	0.574488888888889\\
0.051	0.598142222222222\\
0.054	0.619075555555555\\
0.057	0.639688888888889\\
0.06	0.659413333333333\\
0.063	0.678186666666667\\
0.066	0.695813333333333\\
0.069	0.712853333333333\\
0.072	0.728462222222222\\
0.075	0.743271111111111\\
0.078	0.757866666666667\\
0.081	0.771093333333333\\
0.084	0.783928888888889\\
0.087	0.795546666666667\\
0.09	0.806817777777778\\
0.093	0.81752\\
0.096	0.827306666666667\\
0.099	0.836417777777778\\
0.102	0.845111111111111\\
0.105	0.85248\\
0.108	0.858924444444444\\
0.111	0.865288888888889\\
0.114	0.871066666666667\\
0.117	0.876195555555556\\
0.12	0.881217777777778\\
0.123	0.885937777777778\\
0.126	0.890444444444444\\
0.129	0.894702222222222\\
0.132	0.898711111111111\\
0.135	0.902328888888889\\
0.138	0.905715555555556\\
0.141	0.90904\\
0.144	0.912177777777778\\
0.147	0.914924444444444\\
0.15	0.917484444444444\\
0.153	0.920062222222222\\
0.156	0.922471111111111\\
0.159	0.924933333333333\\
0.162	0.927208888888889\\
0.165	0.929626666666667\\
0.168	0.932195555555555\\
0.171	0.934186666666667\\
0.174	0.936213333333333\\
0.177	0.938311111111111\\
0.18	0.940204444444445\\
0.183	0.942186666666667\\
0.186	0.944106666666667\\
0.189	0.945848888888889\\
0.192	0.9476\\
0.195	0.949386666666667\\
0.198	0.950942222222222\\
0.201	0.952711111111111\\
0.204	0.95448\\
0.207	0.956133333333333\\
0.21	0.957946666666667\\
0.213	0.959813333333333\\
0.216	0.961422222222222\\
0.219	0.963057777777778\\
0.222	0.964622222222222\\
0.225	0.966284444444444\\
0.228	0.967671111111111\\
0.231	0.969182222222222\\
0.234	0.970622222222222\\
0.237	0.972026666666667\\
0.24	0.97336\\
0.243	0.974631111111111\\
0.246	0.975822222222222\\
0.249	0.977004444444444\\
0.252	0.978328888888889\\
0.255	0.979715555555555\\
0.258	0.980897777777778\\
0.261	0.981884444444444\\
0.264	0.982853333333333\\
0.267	0.983866666666667\\
0.27	0.984666666666667\\
0.273	0.985591111111111\\
0.276	0.986604444444445\\
0.279	0.98744\\
0.282	0.988302222222222\\
0.285	0.989253333333333\\
0.288	0.990133333333333\\
0.291	0.99096\\
0.294	0.991635555555555\\
0.297	0.992195555555555\\
0.3	0.992853333333333\\
};
\addlegendentry{ADD3 \cite{add16}} %

\end{axis}
\end{tikzpicture}
\end{figure}

\subsubsection{Results on SCAPE : real and partial data.}
The SCAPE dataset provides meshes aligned to real scans and includes poses different from our training dataset. When applying a network trained directly on our SMPL data, we obtain satisfying performance, namely 3.14cm average Euclidean error. Quantitative comparison of correspondence quality in terms of geodesic error are given in Fig~\ref{fig:quantitative}. We outperform all methods except for Deep Functional Maps~\cite{Litany17}. %
SCAPE also allows evaluation on real partial scans. Quantitatively, the error on these partial meshes is 4.04cm, similar to the performance on the full meshes. Qualitative results are shown in Fig~\ref{fig:SCAPE}.

\subsubsection{Results on TOSCA : robustness to perturbations.}
The TOSCA dataset provides several versions of the same synthetic mesh with different perturbations. We found that our method, still trained only on SMPL or SMAL data, is robust to all perturbations (isometry, noise, shotnoise, holes, micro-holes, topology changes, and sampling), except scale, which can be trivially fixed by normalizing all meshes to have consistent surface area. Examples of representative qualitative results are shown Fig~\ref{fig:TOSCA} and quantitative results are reported in appendix~\cite{appendix}.

\begin{table}[b]
\centering
{
  \begin{tabular}{l|c}
  \hline
  Method & Faust error (cm) \\

  \hline
  \hline
   Without regression & 6.29  \\
 With regression  & { 3.255}   \\
 With regression + Regular Sampling  & { 3.048}   \\
 With regression + Regular Sampling + High-Res template & {\bf 2.878}   \\

  \end{tabular}
  }
      \caption{{\bf Importance of the reconstruction optimization step.} Optimizing the latent feature is key to our results. Regular point sampling for training and high resolution for the nearest neighbor step provide an additional boost. 
     }\label{tab:ablation_reg}
\end{table}

\begin{figure}[t]
\centering
\begin{subfigure}[b]{0.15\linewidth}
\centering
 \includegraphics[height=2.5cm]{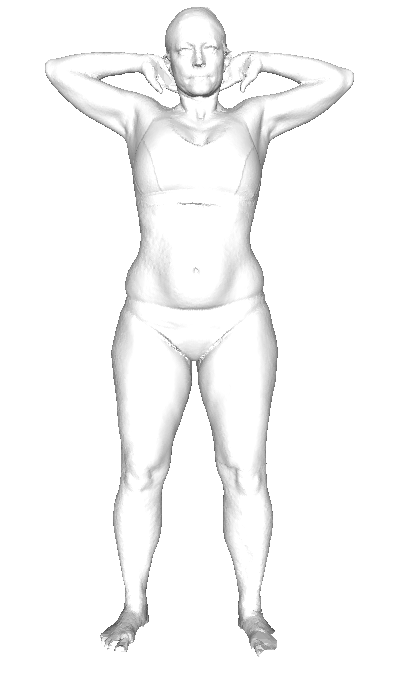}
 \caption{Input }
\end{subfigure}
\begin{subfigure}[b]{0.25\linewidth}
\centering
 \includegraphics[height=2.5cm]{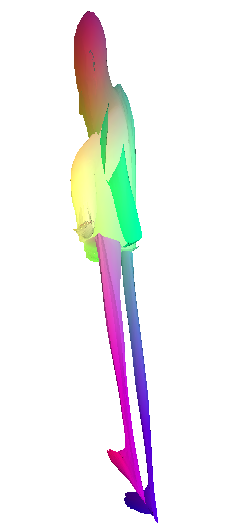}
  \includegraphics[height=2.5cm]{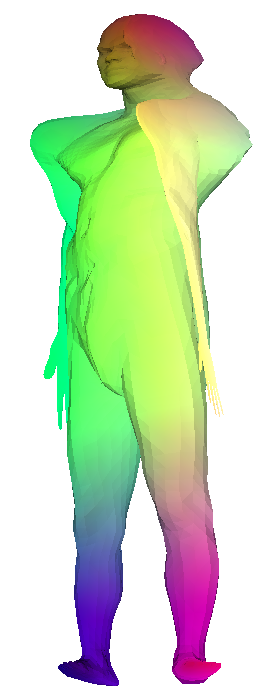}%
 \caption{Random init.}
\end{subfigure}
\begin{subfigure}[b]{0.25\linewidth}
\centering
 \includegraphics[ height=2.5cm]{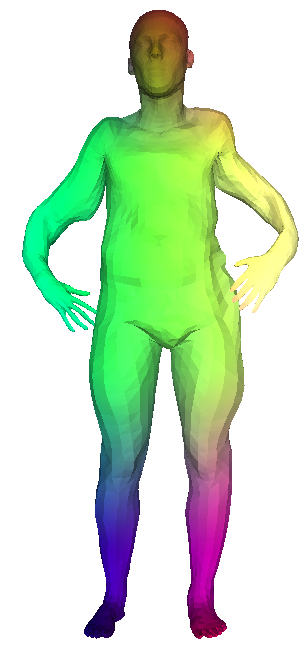}
  \includegraphics[height=2.5cm]{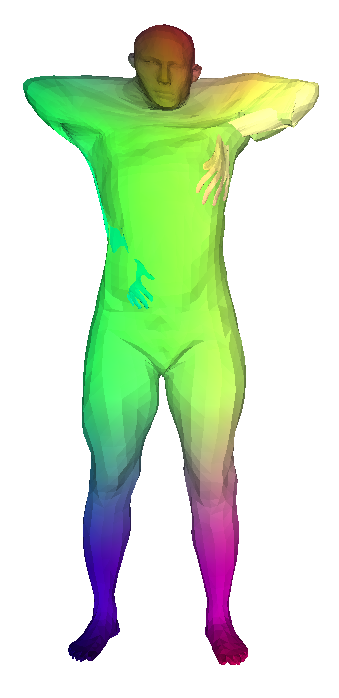}%
\caption{Incorrect init.}
\end{subfigure}
\begin{subfigure}[b]{0.3\linewidth}
\centering
 \includegraphics[height=2.5cm]{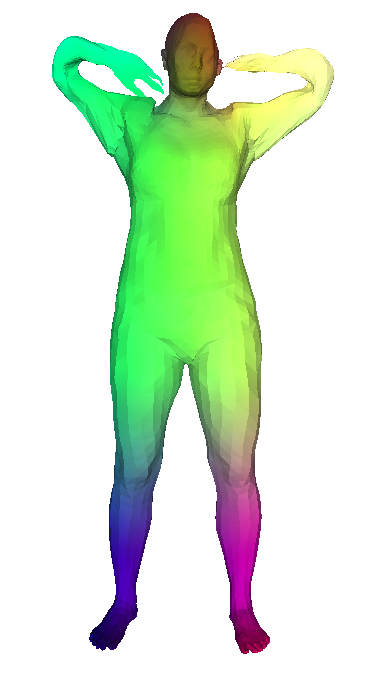} \includegraphics[height=2.5cm]{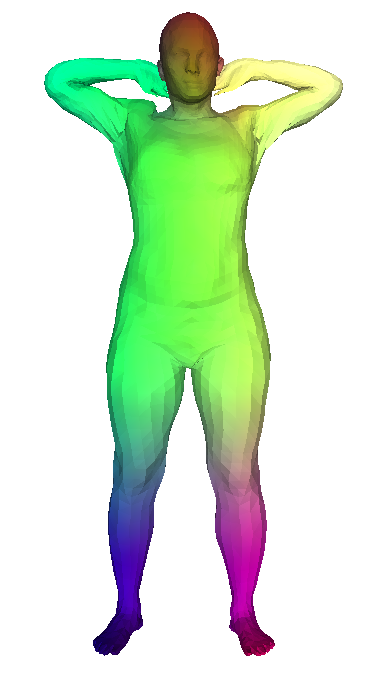}
\caption{Valid init.}
\end{subfigure}

\caption{
{\bf Reconstruction optimization.} The quality of the initialization (i.e. the first step of our algorithm) is crucial for the deformation optimization. For a given target shape (a) and for different initializations (left of (b), (c) and (d)) the figure shows the results of the optimization. If the initialization is random (b) or incorrect (c), the optimization converges to bad local minima. With a reasonable initialization (d) it converges to a shape very close to the target ((d), right). %
}
  \label{fig:regression}
\end{figure}

\subsubsection{Reconstruction optimization.}
Because the nearest neighbors used in the matching step are sensitive to small errors in alignment, the second step of our pipeline which finds the optimal features for reconstruction, is crucial to obtain high quality results.  This optimization however converges to a good optimum only if it is initialized with a reasonable reconstruction, as visualized in Figure \ref{fig:regression}. Since we optimize using Chamfer distance, and not correspondences, we also rely on the fact that the network was trained to generate humans in correspondence and we expect the optimized shape to still be meaningful. %

Table \ref{tab:ablation_reg} reports the associated quantitative results on FAUST-inter. We can see that: (i) optimizing the latent feature to minimize the Chamfer distance between input and output provides a strong boost; (ii) using a better (more uniform) sampling of the shapes when training our network provided a better initialization; (iii) using a high resolution sampling of the template ($\sim$200k vertices) for the nearest-neighbor step provide an additional small boost in performance.

\begin{table}[b]
\centering
{
  \begin{tabular}{l|c}
  \hline
  training data & Faust error (cm) \\

  \hline
  \hline
  FAUST training set & 18.22  \\
  non-augmented synthetic dataset $2\times 10^5$ shapes & 5.63 \\
  augmented synthetic data, $10^3$ shapes & 5.76 \\
  augmented synthetic data, $10^4$ shapes & 4.70 \\
  augmented synthetic data, $2.3 \times 10^5$ shapes & \textbf{3.26}  \\
  \end{tabular}
  }
      \caption{{\bf FAUST-inter results when training on different datasets.} Adding synthetic data reduce the error by a factor of 3, showing its importance. The difference in performance between the basic synthetic dataset and its augmented version is mostly due to failure on specific poses, as in Figure~\ref{fig:bent} .
      }\label{tab:synthetic}
\end{table}
\begin{figure}[t]
\centering
\begin{subfigure}[b]{0.2\linewidth}
 \includegraphics[height=2cm]{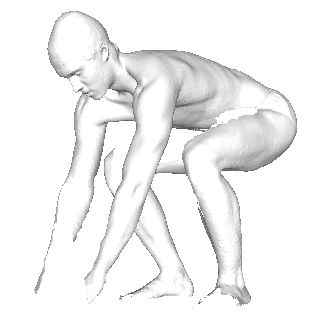}%
\caption{Input}
\end{subfigure}
\begin{subfigure}[b]{0.35\linewidth}
 \includegraphics[height=2cm]{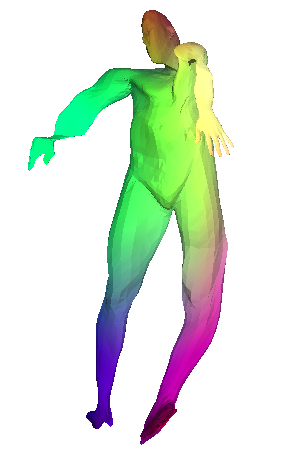}%
 \includegraphics[height=2cm]{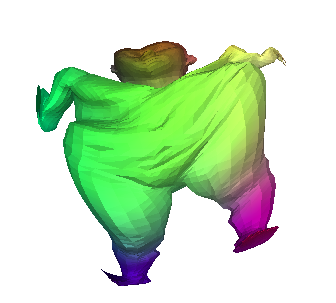}%
\caption{FAUST training data}
\end{subfigure}
\begin{subfigure}[b]{0.35\linewidth} \includegraphics[height=2cm]{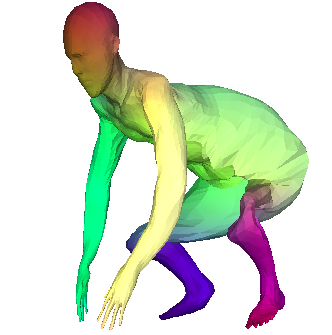}%
 \includegraphics[height=2cm]{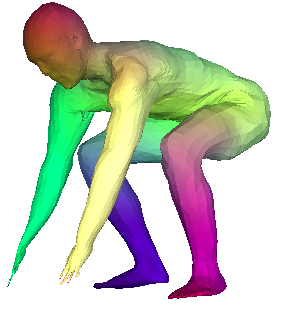}%
 \caption{Augm. synth. training data}
\end{subfigure}
\caption{{\bf Importance of the training data.} For a given target shape (a) reconstructed shapes when the network is trained on FAUST training set (b) and on our augmented synthetic training set (c), before (left) and after (right) the optimization step.}
  \label{fig:synthetic}
\end{figure}
\subsubsection{Necessary amount of training data.}
Training on a large and representative dataset is also crucial for our method. To analyze the effect of training data, we ran our method without re-sampling FAUST points regularly and with a low resolution template %
for different training sets: FAUST training set, $2 \times 10^5$ SURREAL shapes, and $2.3 \times 10^5$, $10^4$ and $10^3$ shapes from our augmented dataset. The quantitative results are reported Table \ref{tab:synthetic} and qualitative results can be seen in Figure \ref{fig:synthetic}. %
The FAUST training set only include 10 different poses and is too small to train our network to generalize. %
Training on many synthetic shapes from the SURREAL dataset~\cite{varol17a} helps overcome this generalization problem. However, if the synthetic dataset does not include any pose close to test poses (such as bent-over humans), the method will fail on these poses (4 test pairs of shapes out of 40). Augmenting the dataset as described in section \ref{sec:data} overcomes this limitation. %
As expected the performance  decreases with the number of training shapes, respectively to 5.76cm and 4.70cm average error on FAUST-inter.%

\begin{table}[b]
\centering
{
  \begin{tabular}{l|c}
  \hline
  Loss & Faust error (cm) \\

  \hline
  \hline
  Chamfer distance, eq. \ref{eqn:atlas_loss} (unsupervised) & 8.727  \\
  Chamfer distance + Regularization, eq. \ref{eqn:training_unsup} (unsupervised)~~ & 4.835  \\
  Correspondences, eq. \ref{eqn:training_sup} (supervised) & \textbf{2.878}  \\
  \end{tabular}
  }
      \caption{Results with and without supervised correspondences. Adding regularization helps the network find a better local minimum in terms of correspondences.
      }\label{tab:unsup}
\end{table}

\subsubsection{Unsupervised correspondences.}
\begin{figure}[t!]
\centering
\begin{subfigure}[b]{0.19\linewidth}
\centering
 \includegraphics[height=2cm]{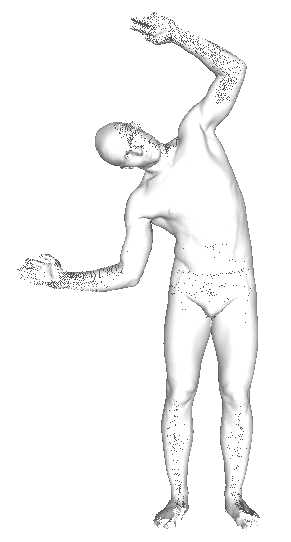}\\
 \includegraphics[height=1.52cm]{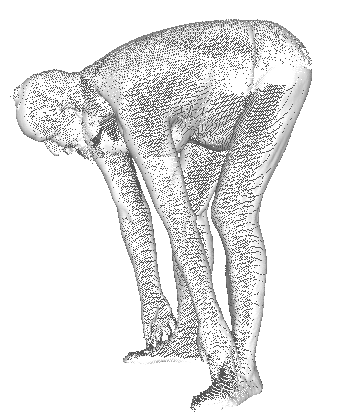}\\
 \includegraphics[height=2cm]{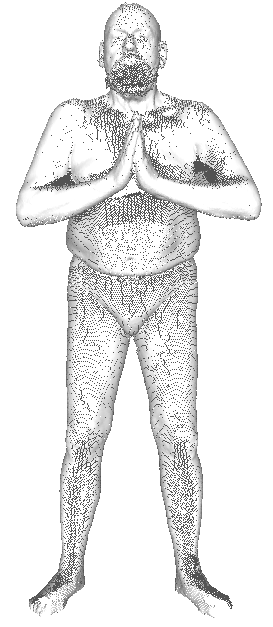}%
\caption{Input (FAUST)}
\end{subfigure}
\begin{subfigure}[b]{0.19\linewidth}
\centering
 \includegraphics[height=2cm]{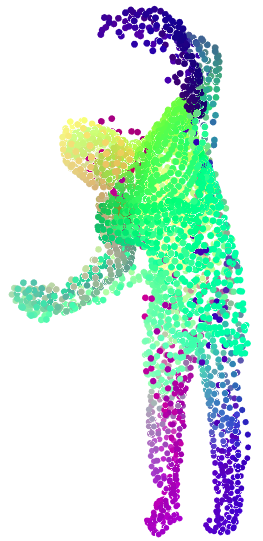}\\
  \includegraphics[height=1.52cm]{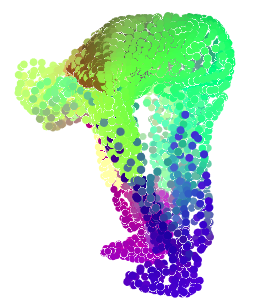}\\
 \includegraphics[height=2cm]{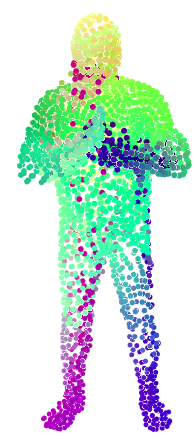}%
 \caption{P.C. after optim.}
\end{subfigure}
\begin{subfigure}[b]{0.19\linewidth}
\centering
 \includegraphics[height=2cm]{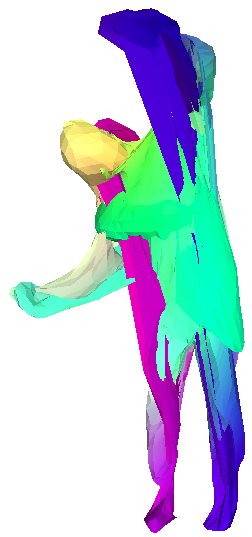}\\
  \includegraphics[height=1.52cm]{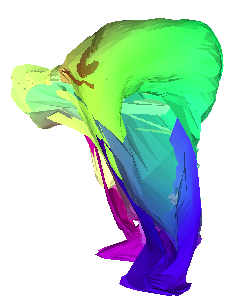}\\
 \includegraphics[height=2cm]{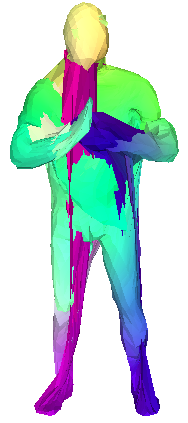}%
 \caption{Mesh after optim.}
\end{subfigure}
\begin{subfigure}[b]{0.19\linewidth}
\centering
 \includegraphics[height=2cm]{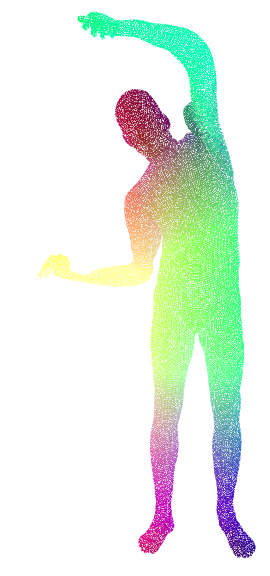}\\
 \includegraphics[height=1.52cm]{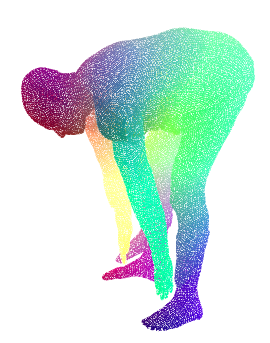}\\
  \includegraphics[height=2cm]{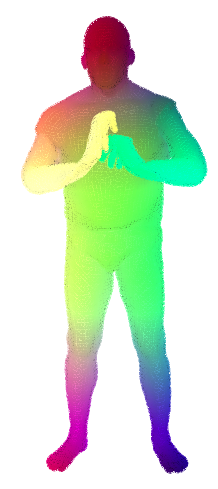}\\
 \caption{P.C. after optim + Regul}
\end{subfigure}
\begin{subfigure}[b]{0.19\linewidth}
\centering
 \includegraphics[height=2cm]{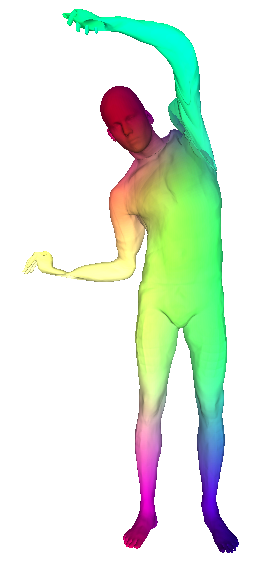}\\
  \includegraphics[height=1.52cm]{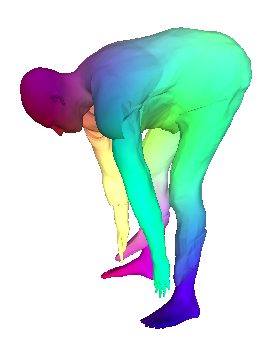}\\
 \includegraphics[height=2cm]{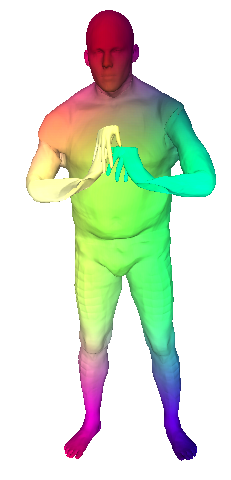}%
 \caption{Mesh after optim + Regul}
\end{subfigure}
\caption{
{\bf Unsupervised correspondences.} %
We visualize for different inputs (a), the point clouds (P.C.) predicted by our %
approach (b,d) and the corresponding meshes (c,e). Note that without regularization, because of the strong distortion, the meshes appear to barely match to the input, while the point clouds are reasonable. On the other hand surface regularization creates reasonable meshes.}
  \label{fig:unsupervised}
  
\end{figure}

We investigate whether our method could be trained without correspondence supervision. We started by simply using the reconstruction loss described in Equation (\ref{eqn:atlas_loss}). One could indeed expect that an optimal way to deform the template into training shapes would respect correspondences. However, we found that the resulting network did not respect correspondences between the template and the input shape, as visualized figure \ref{fig:unsupervised}. However, these results improve with adequate regularization such as the one presented in Equarion (\ref{eqn:training_unsup}), encouraging regularity of the mapping between the template and the reconstruction. %
We trained such a network with the same training data as in the supervised case but {\bf without any correspondence supervision} and obtained a 4.88cm of error on the FAUST-inter data, i.e. similar to Deep Functional Map~\cite{Litany17} which had an error of 4.83 cm. This demonstrates that our method can be efficient even without correspondence supervision. Further details on regularization losses are given in the appendix~\cite{appendix}.

\subsubsection{Rotation invariance}
We handled rotation invariance by rotating the shape and selecting the orientation for which the reconstruction is optimal. As an alternative, we tried to learn a network directly invariant to rotations around the vertical axis. It turned out the performances were slightly worse on FAUST-inter (3.10cm), but still better than the state of the art. We believe this is due to the limited capacity of the network and should be tried with a larger network. However, interestingly, this rotation invariant network seems to have increased robustness and provided slightly better results on SCAPE.%
\section{Conclusion}

We have demonstrated an encoder-decoder deep network architecture that can generate human shape correspondences competitive with state-of-the-art approaches and that uses only simple reconstruction and correspondence losses. Our key insight is to factor the problem into an encoder network that produces a global shape descriptor, and a decoder Shape Deformation Network that uses this global descriptor to map points on a template back to the original geometry. A straightforward regression step uses gradient descent through the Shape Deformation Network to significantly improve the final correspondence quality.

\myparagraph{Acknowledgments.} This work was partly supported by ANR project EnHerit ANR-17-CE23-0008, Labex B\'ezout, and gifts from Adobe to \'Ecole des Ponts. We thank G{\"u}l Varol, Angjoo Kanazawa, and Renaud Marlet for fruitful discussions. 
\bibliographystyle{splncs}
\bibliography{1804}

\begin{thebibliography}{10}
\providecommand{\url}[1]{\texttt{#1}}
\providecommand{\urlprefix}{URL }
\providecommand{\doi}[1]{https://doi.org/#1}

\bibitem{Allen02}
Allen, B., Curless, B., Popovic, Z.: Articulated body deformation from range
  scan data. SIGGRAPH  (2002)

\bibitem{Allen03}
Allen, B., Curless, B., Popovic, Z.: The space of human body shapes:
  reconstruction and parameterization from range scans. SIGGRAPH  (2003)

\bibitem{Allen06}
Allen, B., Curless, B., Popovic, Z.: Learning a correlated model of identity
  and pose-dependent body shape variation for real-time synthesis. Symposium on
  Computer Animation  (2006)

\bibitem{anguelov2005scape}
Anguelov, D., Srinivasan, P., Koller, D., Thrun, S., Rodgers, J., Davis, J.:
  Scape: shape completion and animation of people. ACM transactions on graphics
  (TOG)  \textbf{24}(3),  408--416 (2005)

\bibitem{Aubry11}
Aubry, M., Schlickewei, U., Cremers, D.: The wave kernel signature: A quantum
  mechanical approach to shape analysis. IEEE International Conference on
  Computer Vision (ICCV) - Workshop on Dynamic Shape Capture and Analysis
  (4DMOD)  (2011)

\bibitem{Bogo:CVPR:2014}
Bogo, F., Romero, J., Loper, M., Black, M.J.: {FAUST}: Dataset and evaluation
  for {3D} mesh registration. Proceedings IEEE Conf. on Computer Vision and
  Pattern Recognition (CVPR)  (Jun 2014)

\bibitem{Bosciani16}
Boscaini, D., Masci, J., Rodola, E., Bronstein, M.M.: Learning shape
  correspondence with anisotropic convolutional neural networks. NIPS  (2016)

\bibitem{WFT2015}
Boscaini, D., Masci, J., Melzi, S., Bronstein, M.M., Castellani, U.,
  Vandergheynst, P.: Learning class-specific descriptors for deformable shapes
  using localized spectral convolutional networks. Computer Graphics Forum
  \textbf{34}(5),  13--23 (2015)

\bibitem{add16}
Boscaini, D., Masci, J., Rodol\`a, E., Bronstein, M.M., Cremers, D.:
  Anisotropic diffusion descriptors. Computer Graphics Forum  \textbf{35}(2),
  431--441 (2016)

\bibitem{Bronstein06}
Bronstein, A.M., Bronstein, M.M., Kimmel, R.: Efficient computation of
  isometry-invariant distances between surfaces. SIAM J. Scientific Computing
  (2006)

\bibitem{Bronstein06gmds}
Bronstein, A.M., Bronstein, M.M., Kimmel, R.: Generalized multidimensional
  scaling: a framework for isometry-invariant partial surface matching. Proc.
  National Academy of Sciences (PNAS)  (2006)

\bibitem{bronstein2008numerical}
Bronstein, A.M., Bronstein, M.M., Kimmel, R.: Numerical geometry of non-rigid
  shapes. Springer Science \& Business Media (2008)

\bibitem{Chen15}
Chen, Q., Koltun, V.: Robust nonrigid registration by convex optimization.
  International Conference on Computer Vision (ICCV)  (2015)

\bibitem{Raviv13}
D.Raviv, A.Dubrovina, R.Kimmel: Hierarchical framework for shape
  correspondence. Numerical Mathematics: Theory, Methods and Applications
  (2013)

\bibitem{Ezuz17}
Ezuz, D., Solomon, J., Kim, V.G., Ben-Chen, M.: Gwcnn: A metric alignment layer
  for deep shape analysis. SGP  (2017)

\bibitem{Fan:2017:cvpr}
Fan, H., Su, H., Guibas, L.: A point set generation network for 3{D} object
  reconstruction from a single image. Proceedings of IEEE Conference on
  Computer Vision and Pattern Recognition (CVPR)  (2017)

\bibitem{Girdhar16b}
Girdhar, R., Fouhey, D., Rodriguez, M., Gupta, A.: Learning a predictable and
  generative vector representation for objects. Proceedings of European
  Conference on Computer Vision (ECCV)  (2016)

\bibitem{groueix2018}
Groueix, T., Fisher, M., Kim, V.G., Russell, B., Aubry, M.: {AtlasNet: A
  Papier-M\^ach\'e Approach to Learning 3D Surface Generation}. Proceedings
  IEEE Conf. on Computer Vision and Pattern Recognition (CVPR)  (2018)

\bibitem{appendix}
Groueix, T., Fisher, M., Kim, V.G., Russell, B., Aubry, M.: Supplementary
  material (appendix) for the paper
  \url{https://http://imagine.enpc.fr/~groueixt/3D-CODED/index.html}  (2018)

\bibitem{kanazawa2018learning}
Kanazawa, A., Tulsiani, S., Efros, A.A., Malik, J.: Learning category-specific
  mesh reconstruction from image collections. CoRR  \textbf{abs/1803.07549}
  (2018)

\bibitem{Kim12}
Kim, V.G., Li, W., Mitra, N.J., DiVerdi, S., Funkhouser, T.: {Exploring
  Collections of 3D Models using Fuzzy Correspondences}. Transactions on
  Graphics (Proc. of SIGGRAPH)  \textbf{31}(4) (2012)

\bibitem{Kim11}
Kim, V.G., Lipman, Y., Funkhouser, T.: {Blended Intrinsic Maps}. Transactions
  on Graphics (Proc. of SIGGRAPH)  \textbf{30}(4) (2011)

\bibitem{Lipman09}
Lipman, Y., Funkhouser, T.: Mobius voting for surface correspondence. ACM
  Transactions on Graphics (Proc. SIGGRAPH)  \textbf{28}(3) (2009)

\bibitem{Litany17}
Litany, O., Remez, T., Rodola, E., Bronstein, A.M., Bronstein, M.M.: Deep
  functional maps: Structured prediction for dense shape correspondence. ICCV
  (2017)

\bibitem{Loper15}
Loper, M., Mahmood, N., Romero, J., Pons-Moll, G., Black, M.J.: Smpl: A skinned
  multi-person linear model. SIGGRAPH Asia  (2015)

\bibitem{Maron17}
Maron, H., Galun, M., Aigerman, N., Trope, M., Dym, N., Yumer, E., Kim, V.G.,
  Lipman, Y.: Convolutional neural networks on surfaces via seamless toric
  covers. SIGGRAPH  (2017)

\bibitem{MasBosBroVan15}
Masci, J., Boscaini, D., Bronstein, M.M., Vandergheynst, P.: Geodesic
  convolutional neural networks on riemannian manifolds. Proc. of the IEEE
  International Conference on Computer Vision (ICCV) Workshops pp. 37--45
  (2015)

\bibitem{masci15}
Masci, J., Boscaini, D., Bronstein, M.M., Vandergheynst, P.: Geodesic
  convolutional neural networks on riemannian manifolds. 3dRR  (2015)

\bibitem{Memoli05}
M\'{e}moli, F., Sapiro, S.: A theoretical and computational framework for
  isometry invariant recognition of point cloud data. Foundations of
  Computational Mathematics  (2005)

\bibitem{Meyer01}
Meyer, M., Desbrun, M., Schr, P., Barr, A.: Discrete differential-geometry
  operators for triangulated 2-manifolds. Proceedings of Visualization and
  Mathematics  \textbf{3} (11 2001)

\bibitem{Monti17}
Monti, F., Boscaini, D., Masci, J., Rodola, E., Svoboda, J., Bronstein, M.M.:
  Geometric deep learning on graphs and manifolds using mixture model cnns.
  CVPR  (2017)

\bibitem{Ovsjanikov10}
Ovsjanikov, M., M\'{e}rigot, Q., M\'{e}moli, F., Guibas, L.: One point
  isometric matching with the heat kernel. Computer Graphics Forum (Proc. of
  SGP)  (2010)

\bibitem{Ovsjanikov12}
Ovsjanikov, M., Ben-Chen, M., Solomon, J., Butscher, A., Guibas, L.: Functional
  maps: A flexible representation of maps between shapes. ACM Trans. Graph.
  (2012)

\bibitem{qi2016pointnet}
Qi, C.R., Su, H., Mo, K., Guibas, L.J.: Point{N}et: Deep learning on point sets
  for 3{D} classification and segmentation. Proceedings of IEEE Conference on
  Computer Vision and Pattern Recognition (CVPR)  (2017)

\bibitem{Qi:2017:nips}
Qi, C.R., Yi, L., Su, H., Guibas, L.J.: Point{N}et++: Deep hierarchical feature
  learning on point sets in a metric space. Advances in Neural Information
  Processing Systems (NIPS)  (2017)

\bibitem{Rodola14}
Rodola, E., {Rota Bulo}, S., Windheuser, T., Vestner, M., Cremers, D.: Dense
  non-rigid shape correspondence using random forests. CVPR  (2014)

\bibitem{Sahillioglu11}
Sahillioglu, Y., Yemez, Y.: Coarse-to-fine combinatorial matching for dense
  isometric shape correspondence. Computer Graphics Forum  (2011)

\bibitem{Sinha2017}
Sinha, A., Unmesh, A., Huang, Q., Ramani, K.: Surfnet: Generating 3d shape
  surfaces using deep residual networks. Proceedings of IEEE Conference on
  Computer Vision and Pattern Recognition (CVPR)  (2017)

\bibitem{Sinha2016}
Sinha, A., Bai, J., Ramani, K.: Deep learning 3d shape surfaces using geometry
  images. Proceedings of IEEE Conference on Computer Vision and Pattern
  Recognition (CVPR)  (2016)

\bibitem{Solomon12}
Solomon, J., Nguyen, A., Butscher, A., Ben-Chen, M., Guibas, L.: Soft maps
  between surfaces. SGP  (2012)

\bibitem{Solomon16}
Solomon, J., Peyre, G., Kim, V.G., Sra, S.: Entropic metric alignment for
  correspondence problems. Transactions on Graphics (Proc. of SIGGRAPH)  (2016)

\bibitem{Sorkine}
Sorkine, O.: Differential representations for mesh processing. Comput. Graph.
  Forum  \textbf{25},  789--807 (12 2006)

\bibitem{Sun10}
Sun, J., Ovsjanikov, M., Guibas, L.: A concise and provably informative
  multi-scale signature-based on heat diffusion". Computer Graphics Forum
  (Proc. of SGP)  (2009)

\bibitem{Tombari10}
Tombari, F., Salti, S., Stefano, L.D.: Unique signatures of histograms for
  local surface description. ECCV  (2010)

\bibitem{varol17a}
Varol, G., Romero, J., Martin, X., Mahmood, N., Black, M.J., Laptev, I.,
  Schmid, C.: Learning from synthetic humans. CVPR  (2017)

\bibitem{wei2016dense}
Wei, L., Huang, Q., Ceylan, D., Vouga, E., Li, H.: Dense human body
  correspondences using convolutional networks. Computer Vision and Pattern
  Recognition (CVPR)  (2016)

\bibitem{wu20153d}
Wu, Z., Song, S., Khosla, A., Yu, F., Zhang, L., Tang, X., Xiao, J.: 3d
  shapenets: A deep representation for volumetric shapes. Proceedings of the
  IEEE Conference on Computer Vision and Pattern Recognition pp. 1912--1920
  (2015)

\bibitem{Yang2017FoldingNetPC}
Yang, Y., Feng, C., Shen, Y., Tian, D.: Foldingnet: Point cloud auto-encoder
  via deep grid deformation. CVPR  (2018)

\bibitem{Zuffi15}
Zuffi, S., Black., M.J.: The stitched puppet: A graphical model of 3d human
  shape and pose. Proceedings IEEE Conf. on Computer Vision and Pattern
  Recognition (CVPR)  (2015)

\bibitem{Zuffi:CVPR:2018}
Zuffi, S., Kanazawa, A., Black, M.J.: Lions and tigers and bears: Capturing
  non-rigid, {3D}, articulated shape from images. IEEE Conference on Computer
  Vision and Pattern Recognition (CVPR)  (2018)

\bibitem{Zuffi:CVPR:2017}
Zuffi, S., Kanazawa, A., Jacobs, D., Black, M.J.: {3D} menagerie: Modeling the
  {3D} shape and pose of animals. CVPR  (2017)

\end{thebibliography}

\clearpage
\section{Supplementary}

\subsection{Choice of template}
\begin{figure}
\centering
\begin{subfigure}[b]{0.30\linewidth}
\centering
\includegraphics[height=0.75\linewidth]{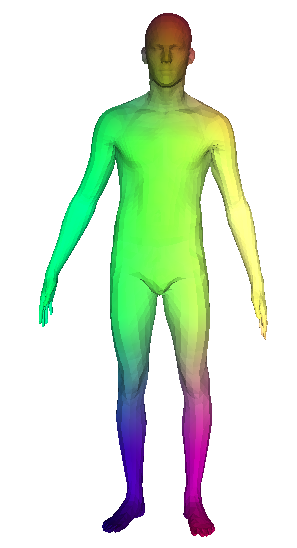}
\caption{``FAUST'' template}
\end{subfigure}
\begin{subfigure}[b]{0.30\linewidth}
\centering
\includegraphics[height=0.75\linewidth]{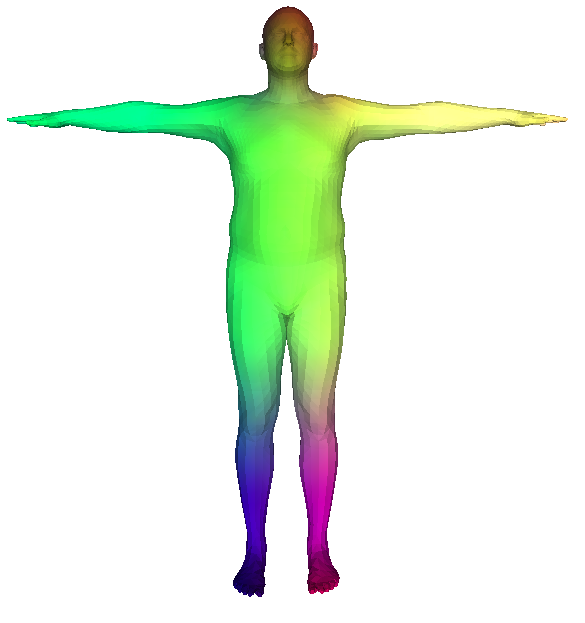}
\caption{``Zero'' template}
\end{subfigure}
\begin{subfigure}[b]{0.30\linewidth}
\centering
 \includegraphics[height=0.75\linewidth]{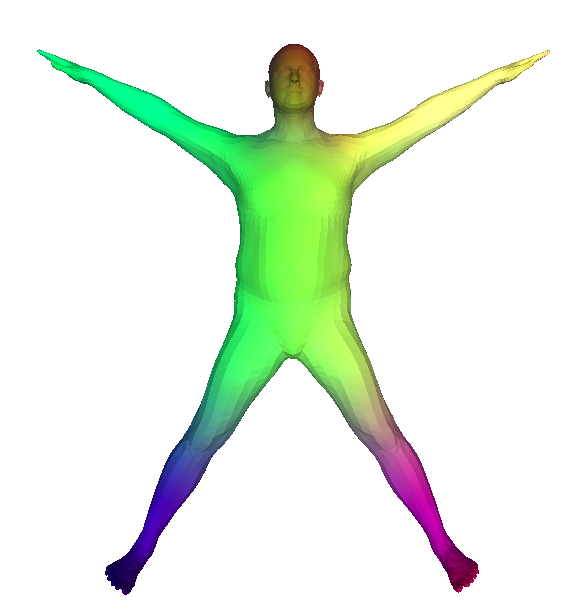}
 \caption{``Separated'' template}
 \end{subfigure}
\caption{
{\bf Shapes for template study.} We evaluate three different template shapes used in our model.
}
  \label{fig:template}
\end{figure}

The template is a critical element for our method. We experimented with three different templates: (i) a ``FAUST'' template associated with SMPL parameters fitted to a body in a neutral pose in the FAUST training set, (ii) a ``zero'' template corresponding to the ``zero'' shape of SMPL, and (iii) a ``separated'' template in which this ``zero'' shape is modified to have the legs better separated and the arms higher. In this experiment,  the points are not sampled regularly on the surface, and a low
resolution template is used. Figure \ref{fig:template} shows the different templates, while table \ref{tab:template} shows quantitative results using the different templates. Interestingly, the best results were obtained with the more ``natural'' template, selected in the ``FAUST'' training dataset, rather than with the templates from simple SMPL parameters, where points from different body parts seem easier to separate.  

\begin{table*}[b]
\centering
{
  \begin{tabular}{l|c}
  \hline
  template 0 & Faust error (cm) \\

  \hline
  \hline
  ``FAUST'' template  &{\bf 3.255}  \\
  ``Zero'' template  & 3.385  \\
  ``Separated'' template  & 3.314  \\
  \end{tabular}
  }
      \caption{
      {\bf Comparison of different template shapes.} We compare different choices for the template shape shown in Figure~\ref{fig:template}. Notice that the neutral ``FAUST'' template performs best out of the three tested shapes.
      }\label{tab:template}
\end{table*}

\subsection{Quantitative results for perturbations on TOSCA}

\begin{figure}[!h]
\centering
\begin{subfigure}[b]{0.3\linewidth}
\centering
 \includegraphics[width=0.45\linewidth]{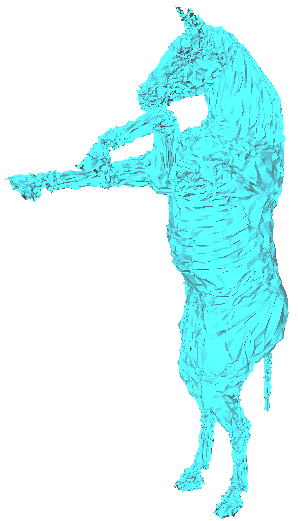}
 \includegraphics[width=0.45\linewidth]{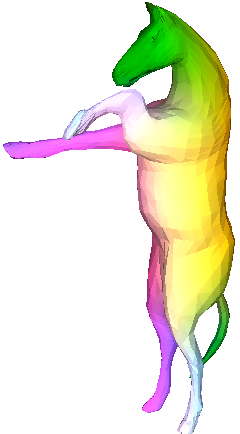}
 \caption{Noise\label{fig:Noise}}
\end{subfigure}~~~~
\begin{subfigure}[b]{0.3\linewidth}
\centering
 \includegraphics[width=0.45\linewidth]{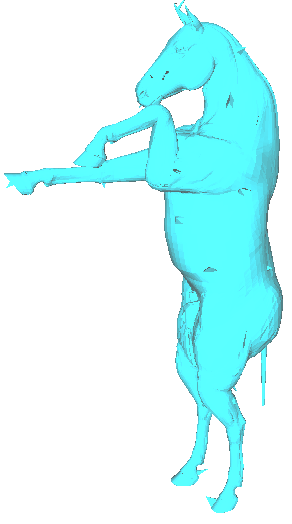}
 \includegraphics[width=0.45\linewidth]{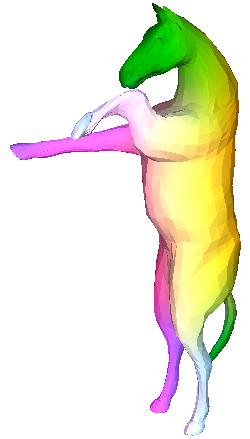}
 \caption{Shotnoise\label{fig:Shotnoise}}
\end{subfigure}~~~~
\begin{subfigure}[b]{0.3\linewidth}
\centering
 \includegraphics[width=0.45\linewidth]{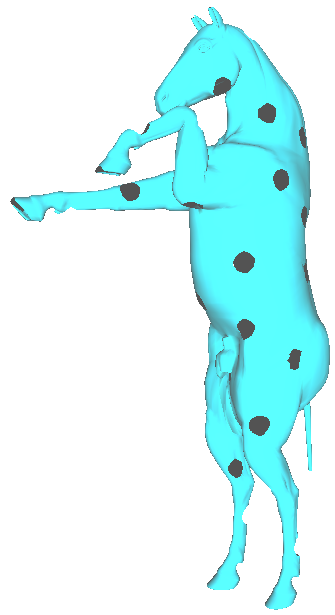}
 \includegraphics[width=0.45\linewidth]{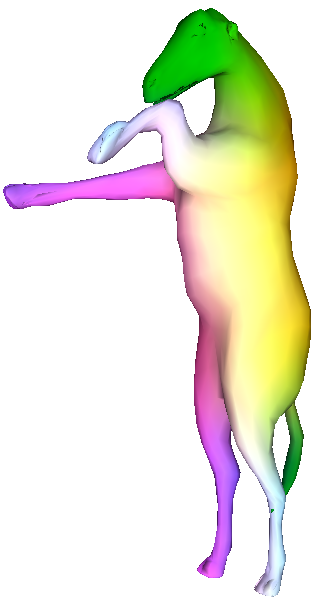}
  \caption{Holes\label{fig:Holes}}
 \end{subfigure}~~

\begin{subfigure}[b]{0.3\linewidth}
\centering
 \includegraphics[width=0.45\linewidth]{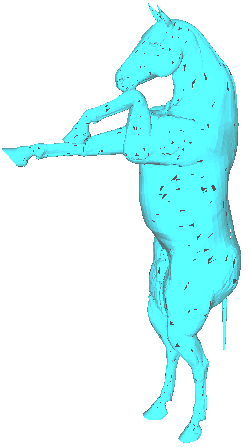}
 \includegraphics[width=0.45\linewidth]{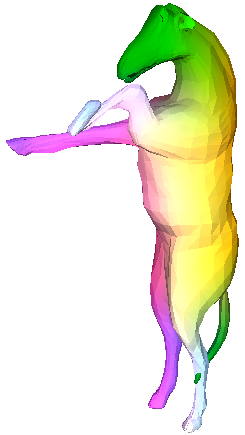}
 \caption{Microholes\label{fig:Microholes}}
\end{subfigure}~~~~
\begin{subfigure}[b]{0.3\linewidth}
\centering
 \includegraphics[width=0.45\linewidth]{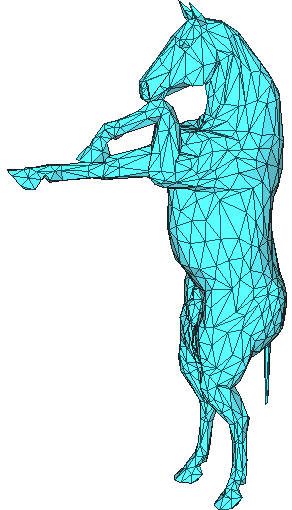}
 \includegraphics[width=0.45\linewidth]{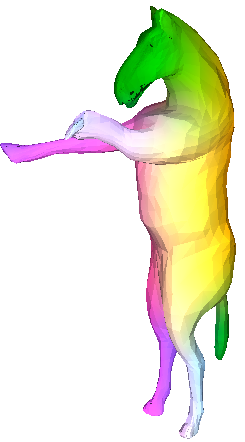}
  \caption{Sampling\label{fig:Sampling}}
 \end{subfigure}~~
\begin{subfigure}[b]{0.3\linewidth}
\centering
 \includegraphics[width=0.45\linewidth]{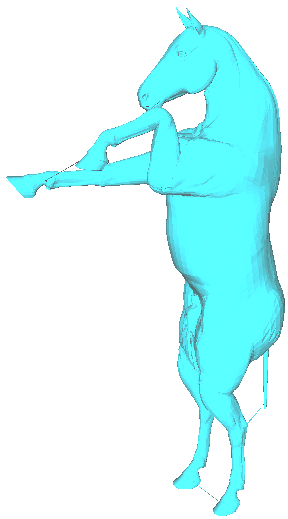}
 \includegraphics[width=0.45\linewidth]{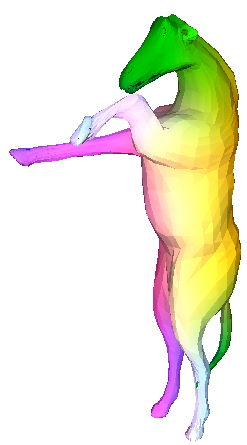}
  \caption{Topology\label{fig:Topology}}
\end{subfigure}

\begin{subfigure}[b]{0.3\linewidth}
\centering
 \includegraphics[width=0.45\linewidth]{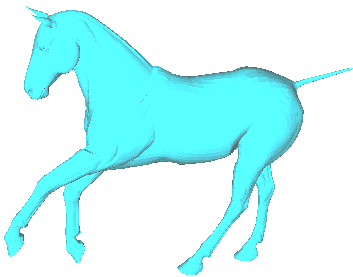}
 \includegraphics[width=0.45\linewidth]{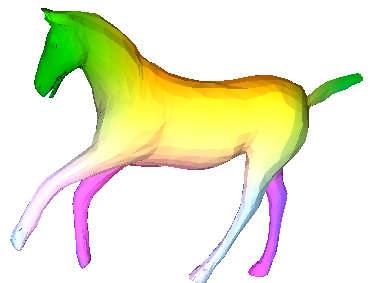}
 \caption{Isometry\label{fig:Isometry}}
\end{subfigure}~~~~
\begin{subfigure}[b]{0.3\linewidth}
\centering
 \includegraphics[width=0.45\linewidth]{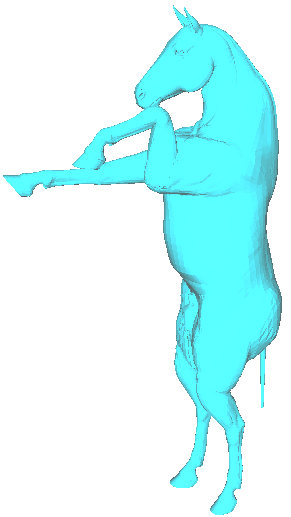}
 \includegraphics[width=0.45\linewidth]{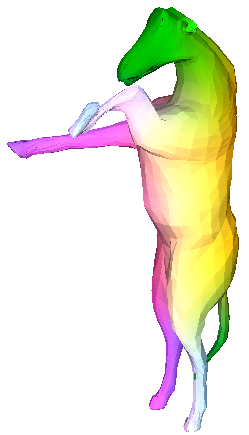}
  \caption{Scale\label{fig:scale}}
 \end{subfigure}~~
\begin{subfigure}[b]{0.3\linewidth}
\centering
 \includegraphics[width=0.45\linewidth]{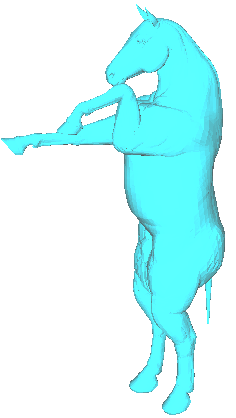}
 \includegraphics[width=0.45\linewidth]{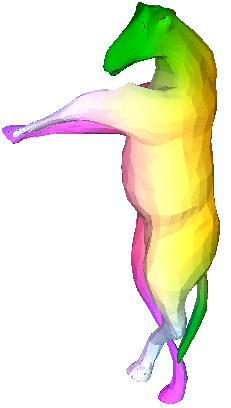}
  \caption{Local scale\label{fig:Local_scale}}
\end{subfigure}

\caption{{\bf Robustness to perturbations on TOSCA for the horse category.} Correspondences are suggested by color. Notice the overall robustness to all perturbations, with small errors on the ears, tail or legs.}
  \label{fig:tosca}
 \end{figure}

We quantitatively evaluate the robustness of our method to perturbation on the TOSCA dataset. This dataset consists of several versions of the same synthetic mesh with different perturbations, specifically: noise, shotnoise, sampling, scale, local scale, topology, holes, microholes, and isometry. We experimented on the horse model. In Table~\ref{tab:tosca} we report quantitative results for each perturbation (with a gradual strength from 1 to 5) and show qualitative reconstruction with correspondences suggested by colors for each category with maximum strength in Figure~\ref{fig:tosca}. We found that we are robust to all categories of noise under study, except for strong variation in sampling (964 points instead of 19948) Surprisingly, adding noise can enhance the quantitative error.

\begin{table}[t]
\centering
\begin{tabular}{l|l|l||l|l|l||l|l|l}
 \hline
 Perturbation &   &  Error (cm) &Perturbation &   &  Error (cm)&Perturbation &   &  Error (cm) \\
  \hline
  \hline
\multirow{4}{*}{Noise} & 1 & 4.58 & \multirow{4}{*}{Scale} & 1 & 4.73 &  \multirow{4}{*}{Holes} & 1 & 4.71 \\
 & 2 & 3.87 &  & 2 & 4.78 &  & 2 & 4.71 \\
 & 3 & 3.93 &  & 3 & 4.66 &  & 3 & 4.72 \\
 & 4 & 3.67 &  & 4 & 4.62 &  & 4 & 4.69 \\
 & 5 & 3.91 &  & 5 & 4.67 &  & 5 & 4.84 \\
 \hline
 \multirow{4}{*}{ShotNoise} & 1 & 4.66 &  \multirow{4}{*}{Local scale} & 1 & 4.18 &  \multirow{4}{*}{Microholes} & 1 & 4.71 \\
 & 2 & 2.64 &  & 2 & 3.65 &  & 2 & 4.72 \\
 & 3 & 3.03 &  & 3 & 3.62 &  & 3 & 4.82 \\
 & 4 & 2.72 &  & 4 & 3.75 &  & 4 & 4.69 \\
 & 5 & 3.00 &  & 5 & 3.56 &  & 5 & 3.53 \\
 \hline
 \multirow{4}{*}{Sampling} & 1 & 4.82 &  \multirow{4}{*}{Topology} & 1 & 3.99 &  \multirow{4}{*}{Isometry} & 1 & 4.72 \\
 & 2 & 4.78 &  & 2 & 4.38 &  & 2 & 4.69 \\
 & 3 & 4.61 &  & 3 & 4.37 &  & 3 & 4.79 \\
 & 4 & 3.72 &  & 4 & 4.31 &  & 4 & 4.85 \\
 & 5 & 9.93 &  & 5 & 7.53 &  & 5 & 4.74 \\
 \hline
 \hline
\end{tabular}
\caption{Quantitative results for perturbations on TOSCA for the horse category}
\label{tab:tosca}
\end{table}

\subsection{Cross-category correspondances on animals}
\begin{figure}[!h]
\centering
 \includegraphics[width=0.20\linewidth]{./figures/horses/scale13.png}
 \includegraphics[width=0.20\linewidth]{./figures/horses/noise03.png}\\
  \includegraphics[width=0.25\linewidth]{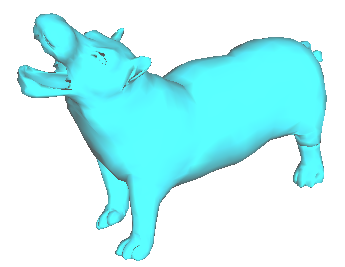}
  \includegraphics[width=0.25\linewidth]{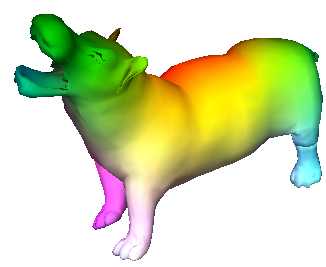}
\caption{{\bf Inter-class correspondences on animals.} Correspondences are suggested by color.}
  \label{fig:hyppos}
 \end{figure}

SMAL synthetic are in correspondences across categories. Hence the template for two different categories are in correspondence and our approach can be trivially extended to get correspondences for animals from different species. Qualitative evidence of this is show in Figure~\ref{fig:hyppos}.

\subsection{Regularization for the unsupervised case}
In the unsupervised case of equation~\ref{eqn:training_unsup}, if the autoencoder is trained using the Chamfer distance alone, it falls into a bad local minimum with high distorsion of the template to reconstruct the input shape.  For example the left foot is propagated on left hand in Figure~\ref{fig:unsupervised}. This distortion is consistent across shapes, so correspondences are still possible, and perform reasonably well with an average error of 8.727cm on the FAUST-inter challenge. However, we expect that by minimizing distorsion in the generated shape, the \textit{Shape Deformation Network} will learn to map an arm to an arm, and a foot to a foot, which will naturally encourage correspondences. We added two regularization losses to achieve this: an edge loss $\lossedges$ and a laplacian loss $\losslap$.
\subsubsection{Edge loss $\lossedges$}
Let $(V,E)$ be the graph of the template and $V^r$ the reconstructed vertices.\\
\begin{equation}
\lossedges(V^r) =\frac{1}{\#E} \cdot \sum_{(i,j)\in E} \mid\frac{\|V^r_i - V^r_j\|}{\|V_i - V_j\|} - 1\mid 
\label{eqn:graph}
\end{equation}
This enforces edges to keep the same length in the template and the generated mesh. We use $\lossedges=0.005$. For instance, if the length of an edge doubles the contribution to the loss is $\lossedges \cdot 1.0 = 0.005 $ which is equivalent (in terms of contribution to the loss function) to a error of placement of 7.1cm. In other words, in terms on loss for the network, it is equivalent to double an edge's length or to misplace a point by 3.2cm. 

\subsubsection{Laplacian loss $\losslap$}
 Similar to Kanazawa et. al. \cite{kanazawa2018learning}, we use the Laplacian regularization.
 The Laplacian matrix $L$ is defined as : 
\begin{equation}
     L_{i,j} = \left\{
                \begin{array}{ll}
                  d_i \text{ if } i=j\\
                  -1 \text{ if } (i,j) \in E\\
                  0 \text{ otherwize }
                \end{array}
              \right.
\end{equation} 
              
\begin{equation}\label{eqn:lv}
\begin{split}
 \left[LV\right]_i & = \sum_{(i,j)\in E} V_i - V_j \\
                &  = d_i \cdot ( V_i - \frac{ \sum_{(i,j)\in E} V_j}{d_i})
 \end{split}
 \end{equation} 

This is an approximation of the following integral as explained in \cite{Sorkine}.
 \begin{equation}
\lim_{\gamma \xrightarrow{}0}\frac{1}{\mid \gamma \mid}\int_{v \in \gamma} (v_i -v)dl(v) = -H(v_i) \cdot n_i
\end{equation} 

where:
\begin{itemize}
    \item $H(v_i)$ is the mean curvature
    \item $n_i$ is the surface normal 
\end{itemize}
We follow \cite{Meyer01} and use cotangent weights in the Laplacian which have been shown to have better geometric discretization.
 \begin{equation}
 \left[L^c V\right]_i = \frac{1}{\Omega_i} \sum_{i\sim j} \frac{1}{2}(cot \alpha_{ij} + cot\beta_{ij})(V_i - V_j)
 \end{equation} 

where :
\begin{itemize}
    \item $\Omega_i$ is the size of the Voronoi cell of i
    \item $ \alpha_{ij}$ and $\beta_{ij}$ denote the two angles opposite of edge (i, j) 
\end{itemize}

Our Laplacian loss is thus written :
\begin{equation}
\losslap(V^g) = \mathds{1}^t \cdot L^c \cdot (V^{template} - V^r)
\end{equation} 
We use $\lambda_{laplace}=0.005$.
In practice we notice that using Laplacian regularization constrains the network to keep sound surfaces.
It may still suffer from error in symmetry and can still invert right and left, and front and back.

\subsection{Asymmetric Chamfer distance}

\begin{figure}
\centering
\begin{subfigure}[b]{0.19\linewidth}
\includegraphics[width=\linewidth]{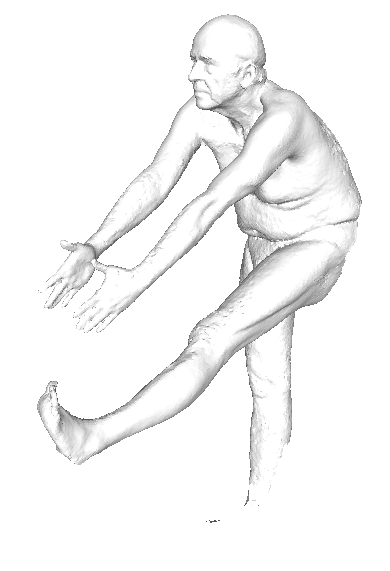}
\caption{target T}
\end{subfigure}
\begin{subfigure}[b]{0.19\linewidth}
\includegraphics[width=\linewidth]{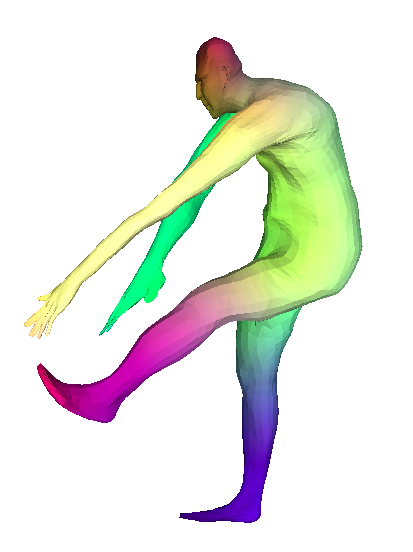}
\caption{Result R}
\end{subfigure}
\begin{subfigure}[b]{0.19\linewidth}
\includegraphics[width=\linewidth]{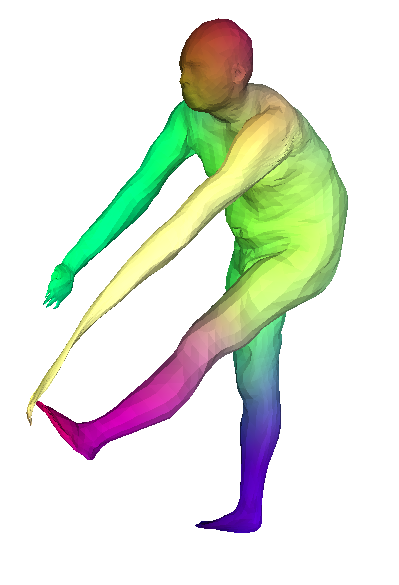}%
  \caption{R attracts T}
\end{subfigure}
\begin{subfigure}[b]{0.19\linewidth}
\includegraphics[width=\linewidth]{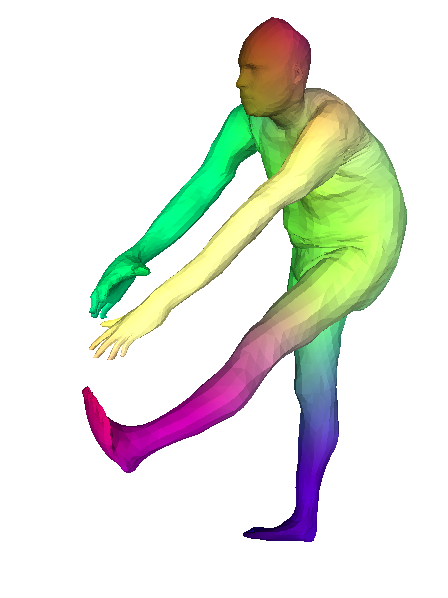}
  \caption{T attracts R}
\end{subfigure}
\begin{subfigure}[b]{0.19\linewidth}
\includegraphics[width=\linewidth]{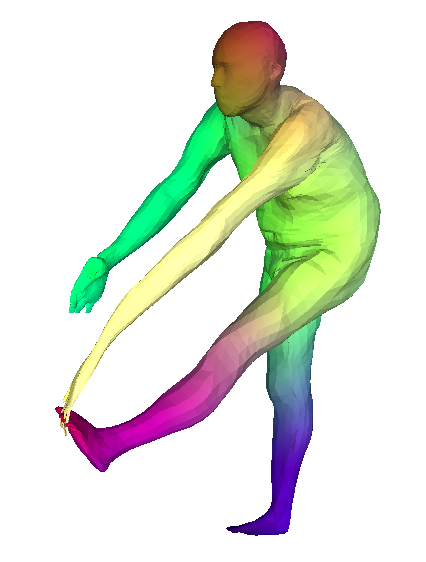}
 \caption{Both ways}
\end{subfigure}
\caption{
{\bf Asymmetric Chamfer loss in reconstruction optimization.} Given an input scan, with holes (a), our network outputs a reconstruction result (b), that can be improved by an optimization step. When the scan has holes, it is better to only consider a loss where the scan attracts the reconstruction (d), rather than using a loss where reconstruction attracts the scan (c), or the Chamfer distance where they attract each other (e).}
  \label{fig:ablation_reg}
\end{figure}

 Figure \ref{fig:ablation_reg} illustrates that optimizing an asymmetric Chamfer distance can in some cases, especially when the 3D scans have holes, produce qualitatively better results. However, Table~\ref{tab:assym} shows that the symmetric version of the Chamfer distance performs better. Investigating how other losses behave, such the Earth-Mover distance loss (also known as Wasserstein loss) behave is left to future work.
 
\begin{table}[h]
\centering
{
  \begin{tabular}{l|c}
  \hline
  Method & Faust error (cm) \\

  \hline
  \hline
   Without regression & 6.29  \\

 With regression, Chamfer asym (R attracts T) & 4.023   \\
 With regression, Chamfer asym (T attracts R) & 3.336   \\
 With regression (both ways)  & { 3.255}   \\

  \end{tabular}
  }
      \caption{{\bf Analysis on the Chamfer distance.} We compare the latent feature search with Chamfer Distance against latent feature searches with asymmetric Chamfer distances. On average, the Chamfer distance (symmetric) performs better  (no regular sampling on the surface, low-resolution template).
     }\label{tab:assym}
\end{table} 

\subsection{Failure cases}

\begin{figure}[h]
\centering
\begin{subfigure}[b]{0.22\linewidth}
\centering
 \includegraphics[height=2cm]{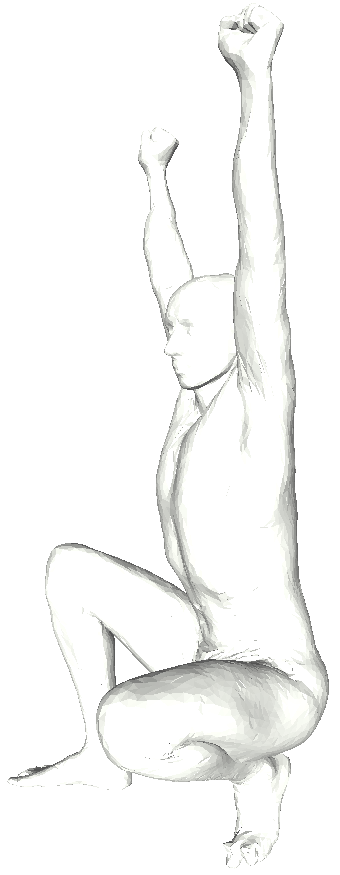}\\
 \includegraphics[height=2cm]{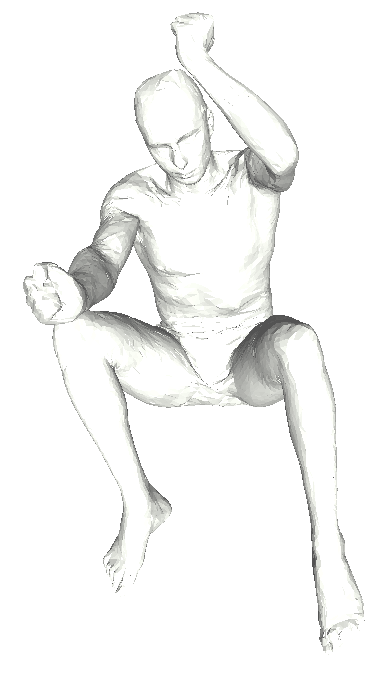}
\caption{Input}
\end{subfigure}
\begin{subfigure}[b]{0.22\linewidth}
\centering
 \includegraphics[height=2cm]{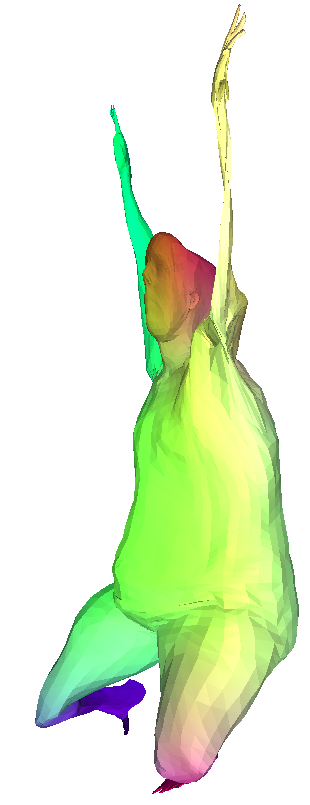}\\
 \includegraphics[height=2cm]{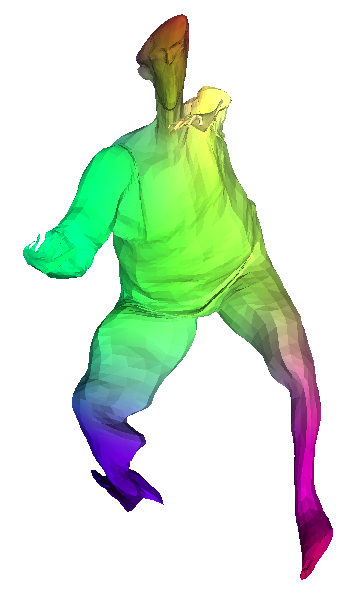}
\caption{Rec. 1}
\end{subfigure}
\begin{subfigure}[b]{0.22\linewidth}
\centering
 \includegraphics[height=2cm]{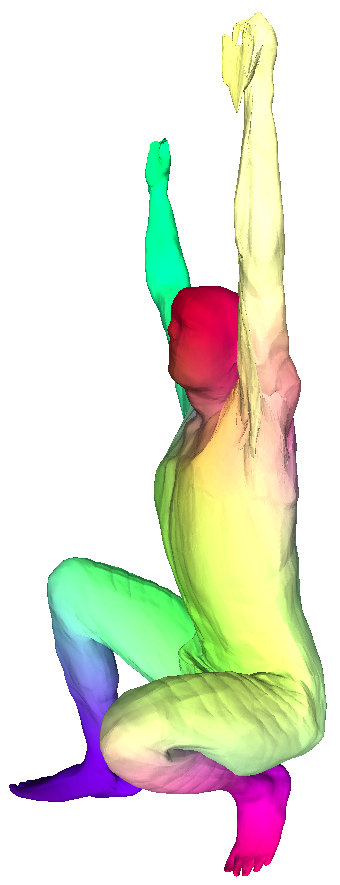}\\
 \includegraphics[height=2cm]{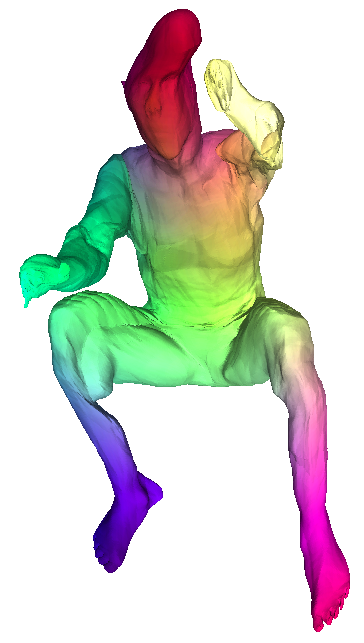}
 \caption{Rec. 2.}
\end{subfigure}
\begin{subfigure}[b]{0.22\linewidth}
\centering
 \includegraphics[height=2cm]{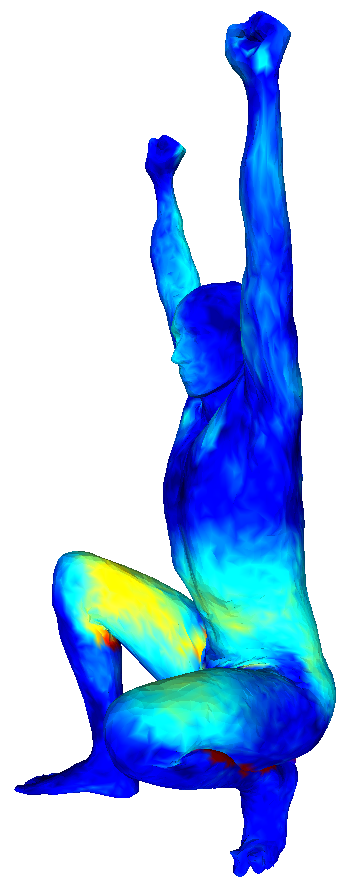}\\
 \includegraphics[height=2cm]{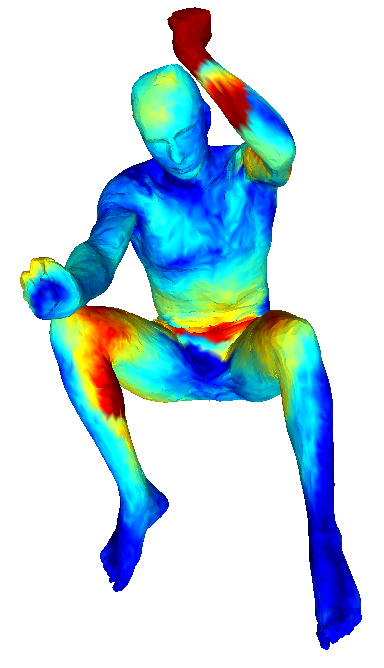}
 \caption{Error.}
\end{subfigure}
\begin{subfigure}[b]{0.05\linewidth}
\centering
 \includegraphics[height=4cm]{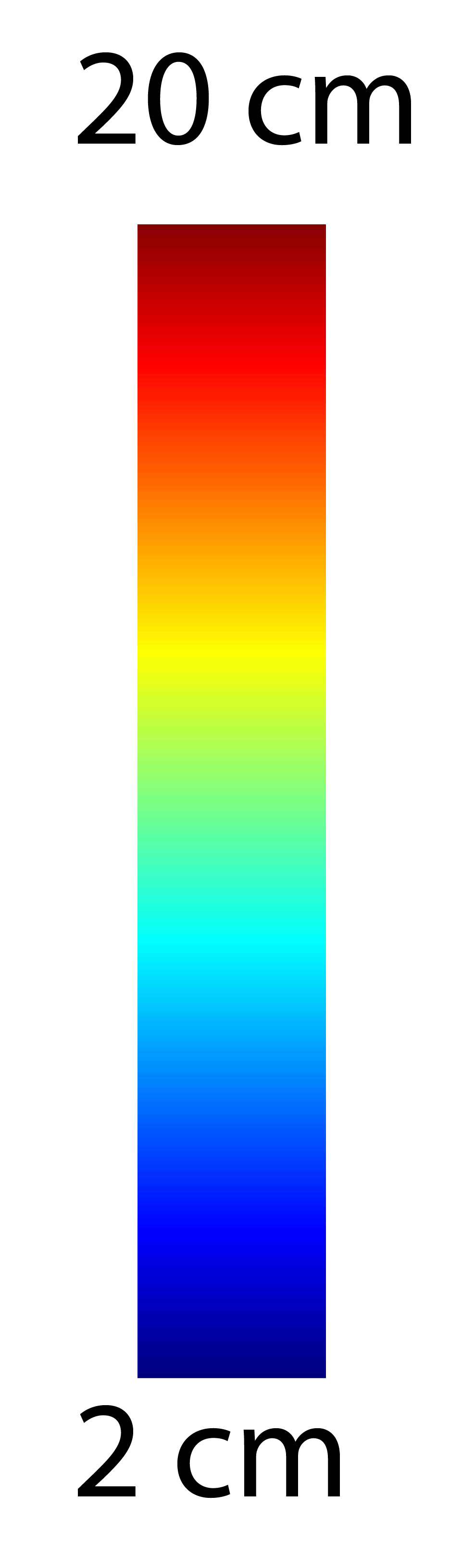}%
\end{subfigure}
\caption{
{\bf Error visualization} Given the input mesh (a), our autoencoder makes an initial reconstruction (b), optimized by a regression step (c). The average in centimeters over each vertex of (a), of the Euclidean distance between its projection and the ground truth, is reported (d). Red vertices have an error higher than 20cm, blue ones lower than 2cm. The largest error are observed in places where the Euclidean distance is small, while the geodesic distance is high, such as touching skin (zoom in on the leg). In such region, the nearest neighbors step is match a vertex in mesh A in a distant (in terms of geodesic distance) vertex in mesh A's reconstruction. High error can also come from a bad reconstruction, such as the head of the second example.}
  \label{fig:error}
\end{figure}

Figure \ref{fig:error} shows the two main sources of error our algorithm faces:
\begin{itemize}
    \item Nearest-neighbor step in overlapping regions failure: a point is matched with the closest point in Euclidean distance but the match is very far in geodesic distance. This could be addressed by adding some regularity in the matches found by the nearest neighbor step. We leave this to future work.
    \item  Failures in reconstruction: in such cases, the initial guess of the autoencoder is just too far away from the input, and the regression step fails.
\end{itemize}

\end{document}